\documentclass[10pt,twocolumn,letterpaper]{article}

\usepackage[pagenumbers]{cvpr} 

\usepackage{graphicx}
\usepackage{amsmath}
\usepackage{amssymb}
\usepackage{booktabs}
\usepackage{kotex}
\usepackage{times}
\usepackage{epsfig}
\usepackage{color}
\usepackage{adjustbox}
\usepackage{algorithm}
\usepackage{algorithmic}
\usepackage{multirow}
\usepackage{lipsum}
\usepackage{paralist}
\usepackage[table]{xcolor}
\usepackage{enumitem}

\newcommand{\sh}[1]{\textcolor{red}{#1}}

\newcommand\norm[1]{\lVert#1\rVert}

\usepackage[pagebackref,breaklinks,colorlinks]{hyperref}

\usepackage[capitalize]{cleveref}
\crefname{section}{Sec.}{Secs.}
\Crefname{section}{Section}{Sections}
\Crefname{table}{Table}{Tables}
\crefname{table}{Tab.}{Tabs.}

\begin{document}

\title{MixNeRF: Modeling a Ray with Mixture Density \\ for Novel View Synthesis from Sparse Inputs}

\author{Seunghyeon Seo~~~
Donghoon Han$^{*}$~~~
Yeonjin Chang$^{*}$~~~
Nojun Kwak
\smallskip
\\
Seoul National Univeristy\\
{\tt\small \{zzzlssh|dhk1349|yjean8315|nojunk\}@snu.ac.kr}
}

\maketitle

\begin{abstract}
   Neural Radiance Field (NeRF) has broken new ground in the novel view synthesis due to its
   simple concept and state-of-the-art quality.
   However, it suffers from severe performance degradation unless trained with a dense set of images with different camera poses, which hinders its practical applications.
   Although previous methods addressing this problem achieved promising results, they relied heavily on the additional training resources, which goes against the philosophy of sparse-input novel-view synthesis pursuing the training efficiency.
   In this work, we propose MixNeRF, an effective training strategy for novel view synthesis from sparse inputs by modeling a ray with a mixture density model.
   Our MixNeRF estimates the joint distribution of RGB colors along the ray samples by modeling it with mixture of distributions.
   We also propose a new task of ray depth estimation as a useful training objective, which is highly correlated with 3D scene geometry.
   Moreover, we remodel the colors with regenerated blending weights based on the estimated ray depth and further improves the robustness for colors and viewpoints.
   Our MixNeRF outperforms other state-of-the-art methods in various standard benchmarks
   with superior efficiency of training and inference.
\end{abstract}

\section{Introduction}
\label{sec:intro}
{\let\thefootnote\relax\footnotetext{{$^{*}$D. Han and Y. Chang equally contributed to this work.}}}

\begin{figure}[!h]
\centering
\vspace{-2mm}
\includegraphics[width=0.9\linewidth]{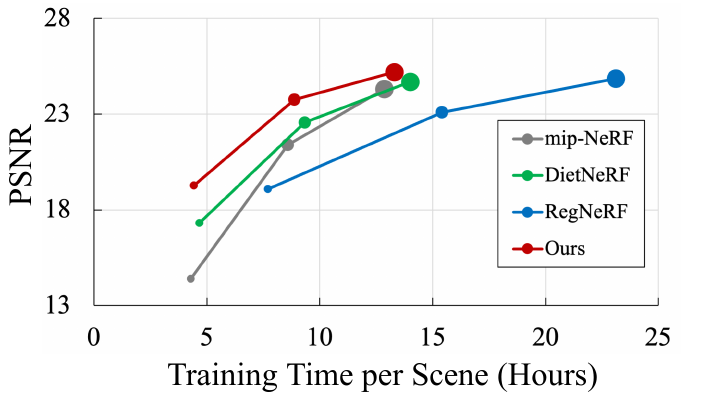}
\vspace{-.3cm}
\caption{\textbf{Comparison with the vanilla mip-NeRF~\cite{barron2021mip} and other regularization methods.} Given the same number of training batch and iterations, our MixNeRF outperforms mip-NeRF and DietNeRF~\cite{jain2021putting} by a large margin with comparable or shorter training time. Compared to RegNeRF~\cite{niemeyer2022regnerf}, ours achieves superior performance with about 42\% shortened training time. The size of the circles are proportional to the number of input views, indicating 3/6/9-view, respectively. More details are provided in \cref{sec:analysis}.}
\vspace{-.3cm}
\label{fig:intro}
\end{figure}

A photo-realistic view synthesis is one of the major research topics in computer vision.
Recently, the coordinate-based neural representation~\cite{chen2019learning, mescheder2019occupancy, michalkiewicz2019implicit, park2019deepsdf} has gained much popularity for the novel view synthesis task.
Among them, Neural Radiance Field (NeRF)~\cite{mildenhall2021nerf}, which models a 3D scene by learning from a dense set of 2D images, enabled high-quality view synthesis with a simple concept and has become the prevailing mainstream.
However, NeRF suffers from severe performance degradation in real-world applications, \eg AR/VR, autonomous driving, and so on, 
where only a sparse set of views are available due to the burdensome task of collecting dense training images.

One of the key factors for  a model's high-quality rendering with limited input views is its robustness in
3D geometry learning, \ie accurate depth estimation for a scene.
There are several works to address this problem
and it can be classified into two major paradigms: \textit{pre-training} and \textit{regularization} approaches.
For the pre-training approach~\cite{yu2021pixelnerf, chen2021mvsnerf, chibane2021stereo, wang2021ibrnet, liu2022neural, jang2021codenerf, li2021mine, rematas2021sharf, trevithick2021grf, johari2022geonerf}, a general 3D geometry is trained by the multi-view images from a large-scale dataset and per-scene finetuning is optionally conducted  in the test time.
Although it has achieved promising results, it still requires the expensive cost for collecting a large-scale dataset across different scenes for pre-training and is not well-generalized for a novel domain in the test time.

Another line of research, the regularization approach~\cite{niemeyer2022regnerf, deng2022depth, roessle2022dense, jain2021putting, kim2022infonerf, xu2022sinnerf}, performs per-scene optimization from scratch by applying regularization to prevent being overfitted from the limited training views.
Most existing methods of this kind depend heavily on the
extra training resources for compensating a lack of supervisory signals, \eg depth-map generation by running SfM~\cite{deng2022depth, roessle2022dense}, unseen ray generation with arbitrary camera poses~\cite{kim2022infonerf, niemeyer2022regnerf}, leveraging external modules to exploit additional features~\cite{jain2021putting, xu2022sinnerf, niemeyer2022regnerf}, or so on.
However,
the additional training data
might not always be available and
the external modules
should be pre-trained with a large-scale dataset.
This is against the philosophy of novel view synthesis from sparse inputs, pursuing training efficiency.

In this work, we propose MixNeRF, an effective regularization approach for novel view synthesis from sparse inputs, modeling the colors along a ray with a mixture density model which represents a complex distribution with a mixture of component distributions.
By exploiting the blending weights as mixing coefficients for our mixture model, we are able to regularize effectively both the colors and the densities of the samples along a ray.
Furthermore, we propose a new auxiliary task of ray depth estimation for learning the 3D geometry which is crucial for the rendering quality.
Since the estimated 3D geometry is highly correlated with the scene depth estimation, our proposed training objective acts as a useful supervisory signal.
Finally, we regenerate the blending weights based on the estimated ray depth and remodel a ray.
Since the estimated depth is not exactly the same, but nearly identical to the ground truth, it can play a role of pseudo geometry for adjacent points of the sample, like an unseen viewpoint.
By remodeling the samples with the mixing coefficients based on the regenerated blending weights, we can further improve the robustness for shift of colors and viewpoints.
Our main contributions are summarized as follows:
\begin{itemize}[noitemsep,topsep=0pt,parsep=0pt,partopsep=0pt, leftmargin=*] 
    \item Our method estimates the joint distribution of RGB color values along the ray samples by a mixture of distributions, learning the 3D geometry successfully with sparse views.
    \item We propose a ray depth estimation as an effective auxiliary task for few-shot novel view synthesis, playing a role of useful training objective.
    \item We use the regenerated blending weights based on the estimated ray depths for improving the robustness
    with negligible extra training cost.
    \item Our MixNeRF outperforms other state-of-the-art methods in the different standard benchmarks,
    showing much improved training and inference efficiency.
\end{itemize}

\section{Related Works}
\subsection{Neural Scene Representations}
Recently, coordinate-based neural representations~\cite{chen2019learning, mescheder2019occupancy, michalkiewicz2019implicit, park2019deepsdf} have gained a lot of popularity in the field of neural scene rendering~\cite{barron2021mip, bergman2021fast, gao2020portrait, jiang2020sdfdiff, kellnhofer2021neural, mildenhall2021nerf, liu2020neural, martin2021nerf}.
Among them, Neural Radiance Fields (NeRF)~\cite{mildenhall2021nerf} have broken new ground in the novel view synthesis research due to its wide possibility with the simple concept and state-of-the-art quality.
Since NeRF, several works have been followed to ameliorate its drawbacks and improve the performance.
Mip-NeRF~\cite{barron2021mip} tackled the problem of aliasing in NeRF by introducing cone tracing method.
Ref-NeRF~\cite{verbin2022ref} reparameterized NeRF from the view-dependent outgoing radiance to reflected radiance, leading to significant improvement for specular reflections.

However, these methods suffer from severe performance degradation unless trained with a set of dense images with different camera poses, which hinders their practical applications.
In this work, we address the sparse input scenario which is closer to the real-world condition.
We are able to perform high-quality view synthesis from sparse inputs by modeling a ray with a mixture density model and improve both the training and the inference efficiency.

\begin{figure*}[t]
\centering
\includegraphics[width=1.0\linewidth]{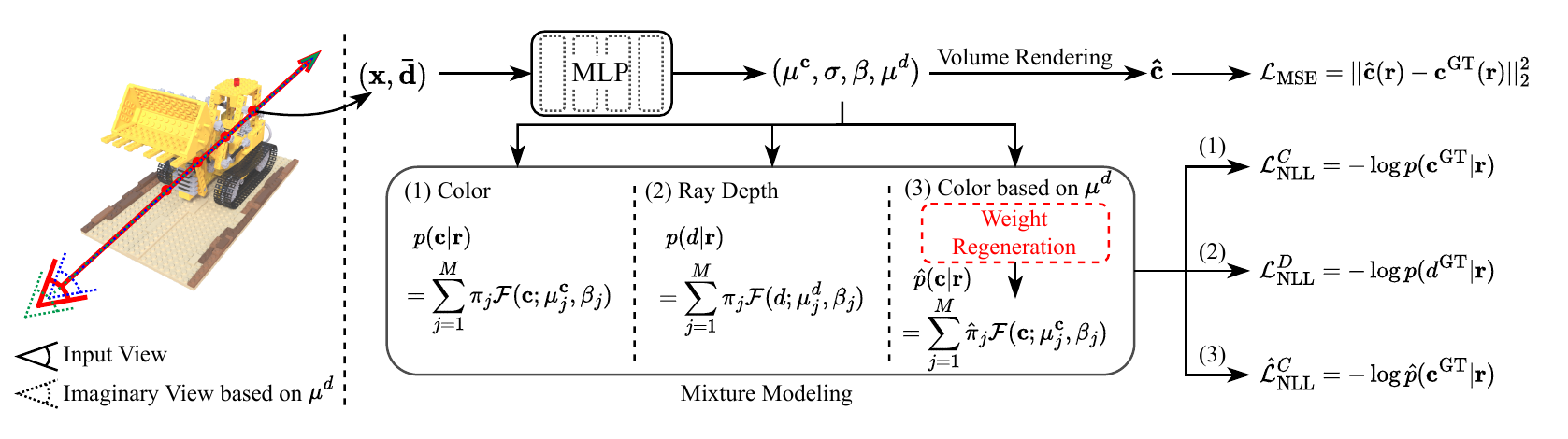}
\vspace{-.85cm}
\caption{\textbf{Overview of MixNeRF.} Our method models the color and depth of a ray with a mixture of distributions, and remodels the color based on the estimated ray depth $\mu^{d}$.
The dotted rays indicate the imaginary views corresponding to $\mu^{d}$.
See \cref{sec:method} for more details.}
\vspace{-.3cm}
\label{fig:overview}
\end{figure*}

\subsection{Sparse Input Novel View Synthesis}
One of the fundamental causes of performance degradation is the lack of 3D geometry information from training images, resulting in an inaccurate depth estimation.
There are two major paradigms to tackle this problem in the novel view synthesis from sparse inputs: \textit{pre-training} and \textit{regularization} approaches.
The former approach~\cite{yu2021pixelnerf, chen2021mvsnerf, chibane2021stereo, wang2021ibrnet, liu2022neural, jang2021codenerf, li2021mine, rematas2021sharf, trevithick2021grf, johari2022geonerf} provides prior knowledges to conditional models through pre-training.
The image features extracted by a CNN feature extractor~\cite{yu2021pixelnerf, chibane2021stereo} or a 3D cost volume obtained by image warping~\cite{chen2021mvsnerf, johari2022geonerf} are used for training a generalizable model.
Although they achieved promising performances under the sparse input setting, a large-scale dataset of multi-view images with different scenes is required for pre-training, which is burdensome to collect.
Furthermore, despite the lengthy pre-training phase, most of these methods require additional test-time fine-tuning and are apt to suffer from quality degradation on different data domains.

The regularization approach~\cite{niemeyer2022regnerf, deng2022depth, roessle2022dense, jain2021putting, kim2022infonerf, xu2022sinnerf} introduces extra supervision to regularize the color and the geometry without an expensive pre-training process.
Additional training resources, \eg external modules such as CLIP~\cite{jain2021putting} or a pre-trained normalizing flow model~\cite{dinh2016density}, extra depth inputs obtained by running structure-from-motion (SfM), and additional rays of unseen viewpoints, are often used to provide abundant supervisory signals.
However, the existing methods are overly dependent on the extra training resources which might not always be available, hampering data/time efficiency.
Moreover, it goes against the philosophy of the sparse-input novel-view synthesis which pursues the training efficiency.

Our proposed method requires neither an external module nor an additional inference of extra supervisory signals, such as additional depth inputs or pre-generated rays from unobserved viewpoints, resulting in a more efficient training framework.

\subsection{Mixture Density Model}
There exists a line of research utilizing a mixture density model in different tasks of computer vision~\cite{truong2021learning, yoo2021training, yoo2022sparse, li2019generating, choi2022md3d, tosi2021smd}.
Among 3D vision tasks, Tosi \etal~\cite{tosi2021smd} proposed a novel stereo-matching framework, SMD-Nets, tackling the over-smoothing problem of output representations by leveraging a mixture density network~\cite{bishop1994mixture}.
Choi \etal~\cite{choi2022md3d} reformulated 3D bounding box regression as a density estimation problem using a Gaussian Mixture Model (GMM), achieving a more efficient 3D object detection framework with few heuristic design factors.

Although the mixture density model shows a great potential in 3D vision tasks, there has not been an attempt to utilize it in the NeRF framework for novel view synthesis.
Our MixNeRF is able to learn the 3D geometry successfully, which is a critical factor for rendering quality under the sparse input setting, by modeling a ray with a mixture of distributions.

\section{Method}
\label{sec:method}
In this work, we propose a novel training framework of neural radiance fields for novel view synthesis from sparse inputs.
We build our MixNeRF upon mip-NeRF~\cite{barron2021mip} which uses a multiscale scene representation (\cref{subsec:preliminary}).
Moreover, we leverage the mixture density model framework to learn 3D geometry efficiently.
More specifically, we model the colors of samples along a ray by a mixture of Laplace distributions with the predicted weights as mixing coefficients, which contributes to learning a scene's geometry effectively with limited input views (\cref{subsec:modeling}).
Furthermore, we estimate the depths of input rays as an auxiliary task and reuse it for producing blending weights once again as supplemental training resources, which enables robust rendering from unseen viewpoints with little additional burden for training (\cref{subsec:auxray}).
In the training phase, our MixNeRF is not only trained to minimize the mean squared error (MSE) between predictions and GT colors, but also to maximize the likelihood of colors and depths for each ray (\cref{subsec:total_loss}).
\cref{fig:overview} demonstrates an overview of our MixNeRF.

\subsection{Preliminary: Neural Radiance Field}
\label{subsec:preliminary}
NeRF~\cite{mildenhall2021nerf} represents a 3D scene with a continuous function, where a neural network $f(\cdot, \cdot)$ consisting of an MLP maps a 3D location $\mathbf{x} = (x, y, z)$ and viewing direction $(\theta, \phi)$, which is expressed as a 3D Cartesian unit vector $\mathbf{\bar{d}}$ in practice, along rays to colors $\mathbf{c} = (r, g, b)$ and volume density $\sigma$:
\begin{equation}
  f(\gamma(\mathbf{x}), \gamma(\bar{\mathbf{d}})) \to (\mathbf{c}, \sigma),
  \label{eq:nerf}
\end{equation}
where $\gamma(\cdot)$ indicates the positional encoding applied to the inputs $(\mathbf{x}, \bar{\mathbf{d}})$.
Following the volume rendering theory~\cite{max1995optical}, a pixel on an image is rendered by alpha compositing the colors and densities along the ray $\mathbf{r}(t) = \mathbf{o} + t\mathbf{d}$ cast from the camera origin $\mathbf{o}$, where $\mathbf{d}$ is the unnormalized direction vector, \ie $\mathbf{d} = \norm{\mathbf{d}}_2 \cdot \bar{\mathbf{d}}$.
The volume rendering
integrals are approximated by the quadrature rule in practice~\cite{mildenhall2021nerf} as follows:
\begin{equation}
\begin{split}
\mathbf{\hat{c}(r)} = \sum_{i=1}^{N}T_i(1-\exp(-\sigma_i\delta_i))\mathbf{c}_i , \\
  \text{where} \quad T_i = \exp ( -\sum_{j=1}^{i-1}\sigma_j\delta_j ).
  \label{eq:alpha_composition}
\end{split}
\end{equation}
Note that $N$ and $\delta_i = \norm{\mathbf{d}}_{2} \cdot (t_{i+1} - t_{i})$ denote the number of samples and the interval between the $i$-th sample and its adjacent one, respectively.
To improve rendering efficiency, the two-stage hierarchical sampling is performed: \textit{coarse} and \textit{fine} stage.
The points are sampled uniformly
along a ray in the coarse stage, and then more informed samples are generated in the fine stage based on the density estimated from the coarse stage.
Finally, the radiance field is optimized by minimizing the MSE between the rendered color and ground truth color over the input images:
\begin{equation}
  \mathcal{L}_\text{MSE} = \sum_{\mathbf{r} \in \mathcal{R}} || \mathbf{\hat{c}(r)} - \mathbf{c^\text{GT}(r)} ||^2_2\,,
  \label{eq:mse}
\end{equation}
where $\mathcal{R}$ indicates a set of input rays.

Following RegNeRF~\cite{niemeyer2022regnerf}, we adopt the mip-NeRF~\cite{barron2021mip} representation for our MixNeRF.
Mip-NeRF effectively alleviates the aliasing problem of NeRF by introducing a cone tracing method and an integrated positional encoding.

\subsection{Modeling a Ray with Mixture Density Model}
\label{subsec:modeling}
Given a set of  input rays $\mathcal{R} = \{\mathbf{r}_1, \cdots, \mathbf{r}_K\}$ on training images with the ground truths $G = \{G_1, \cdots, G_K\}$ for each of $K$ pixels, the $i$-th ground truth $G_i$ consists of the RGB color values $\mathbf{c}_i^\text{GT} = \{r_i, g_i, b_i\}$ and the unnormalized 3D ray vector $\mathbf{d}_i^\text{GT}$, \ie $G_i \triangleq \{\mathbf{c}_i^\text{GT}, \mathbf{d}_i^\text{GT} \}$.
Note that $\mathbf{d}^\text{GT}$ is the direction vector corresponding to $t=1$ from the camera center.
First, our MixNeRF estimates the distribution of the RGB color values $\mathbf{c}_i$ along the samples of the ray $\mathbf{r}_i$ on a pixel with a mixture model, which is derived from a weighted combination of component distributions.
As shown in \cref{fig:overview}, in our model, $(\mathbf{c}, \sigma)$, the conventional outputs of NeRF for each sampled point $\mathbf{r}(t)$, are used as a location parameter $\mu^\mathbf{c}$ and to compute a mixing coefficient $\pi$, respectively. In addition to these, a scale parameter $\beta = \{\beta^r, \beta^g, \beta^b\}$
is also estimated in our model.

We assume that every element of $\mathbf{c}_i$ is independent of each other to simplify our mixture model formulation.
Therefore, the $j$-th component's probability density function (pdf) corresponding to the $j$-th sampled point for the $i$-th ray $\mathbf{r}_i$
is as follows:
\begin{align}
\begin{split}
  \mathcal{F}(\mathbf{c}; \mu^\mathbf{c}_{ij}, \beta_{ij}) &= \prod_{c \in \{r, g, b\}} \mathcal{F}(c; \mu^{c}_{ij}, \beta^{c}_{ij}) \\
  &= \prod_{c \in \{r, g, b\}} \frac{1}{2\beta^{c}_{ij}} \exp\left(-\frac{|c -\mu^{c}_{ij}|}{\beta^{c}_{ij}}\right),
  \label{eq:laplace_pdf}
\end{split}
\end{align}
where $\mathcal{F}$ denotes the Laplacian pdf.
The pdf of our mixture model formed by the component distributions above is defined as:
\begin{equation}
p(\mathbf{c}|\mathbf{r}_i) = \sum_{j=1}^{M} \pi_{ij}\mathcal{F}(\mathbf{c}; \mu_{ij}^\mathbf{c}, \beta_{ij}),
\label{eq:mixture_pdf}
\end{equation}
where
$M$ denotes the
number of mixture components which is the same as the number of samples along a ray.
The mixture coefficient $\pi_{ij}$ is derived from the density output $\sigma_{ij}$ as follows:
\begin{equation}
\pi_{j} = \frac{w_j}{\sum_{m=1}^{M} w_m} = \frac{T_j(1-\exp(-\sigma_{j}\delta_{j}))}{\sum_{m=1}^{M} T_m(1-\exp(-\sigma_{m}\delta_{m}))}.
\label{eq:mix_coef}
\end{equation}
Note that we omitted ray index $i$ for simplicity.
Here, $w_{j}$ and $\delta_{j}$ indicate the weight for the alpha compositing and the sample interval, respectively.
Since the mixture components corresponding to the samples with higher weights, which contribute more to the alpha composition of the color than other samples, are likely to have higher $\pi$, we use the normalized weight as a mixing coefficient $\pi$ so that $\sum_{j=1}^{M} \pi_{j} = 1$.

The concept of a mixture model corresponds to that of alpha compositing in that a complex multimodal distribution is able to be represented by the weighted combination of component distributions with mixing coefficients $\pi$, like a pixel value derived from the weighted combination of estimated RGB values along ray samples with blending weights $w$.
Motivated by this conceptual similarity, we are able to model a ray with a mixture of distributions successfully without any heuristic factors.
The mixing coefficients derived from the blending weights provide effective supervisory signals toward the densities, which are the core factor for successfully learning 3D scene geometry with limited input views.

\begin{figure*}[!t]
\centering
\begin{tabular}{p{0.22\textwidth}p{0.17\textwidth}p{0.17\textwidth}p{0.16\textwidth}p{0.155\textwidth}}
     \centering\scriptsize & \centering\scriptsize mip-NeRF~\cite{barron2021mip} & \centering\scriptsize RegNeRF~\cite{niemeyer2022regnerf} & \centering\scriptsize  MixNeRF (Ours) & \centering\scriptsize mip-NeRF~\cite{barron2021mip} (All-view)
    \end{tabular}
\centering
\includegraphics[width=1.0\linewidth]{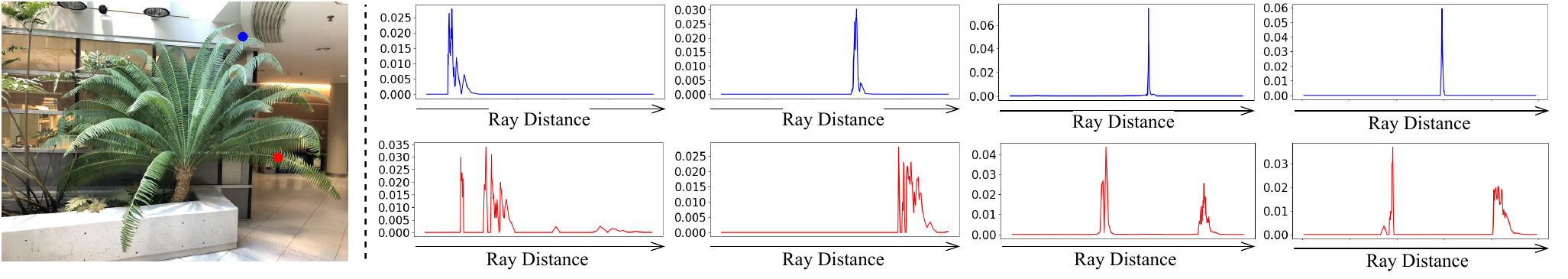}
\vspace{-.7cm}
\caption{\textbf{Comparison of blending weight distributions.}
Compared to the baselines, ours estimates the modes of weight distributions more accurately, leading to the precise 3D geometry.
The ideal distributions on the blue and red points are unimodal and bimodal, respectively.
}
\vspace{-.3cm}
\label{fig:weight_exp}
\end{figure*}

\subsection{Depth Estimation by Mixture Density Model}
\label{subsec:auxray}

We propose a scene's depth estimation 
as an effective auxiliary task for training our MixNeRF with sparse inputs.
As demonstrated in \cref{fig:overview}, our MixNeRF estimates  
$d$, the ray's depth, which is defined as the length of the unnormalized ray direction vector $\mathbf{d}$, \ie $d \triangleq \norm{\mathbf{d}}$, 
along the ray samples.
The ground truth $G_i$ contains the ray direction values $\mathbf{d}_i$ as well, which are used in the form of 3D Cartesian unit vectors $\mathbf{\bar{d}}_i = \mathbf{d}_i / \norm{\mathbf{d}_i}_2$ as an input viewing direction in practice.
Like the RGB color values, the depths for each ray are modeled by our mixture model consisting of the Laplace distributions with the same scale parameters $\beta$ and mixing coefficients $\pi$ used above.
The pdf of our mixture model for the depth of the $i$-th ray is as follows:
\begin{equation}
p(d|\mathbf{r}_i) = \sum_{j=1}^{M} \pi_{ij}\mathcal{F}(d; \mu^d_{ij}, \beta_{ij}).
\label{eq:dir_mixture_pdf}
\end{equation}
Since the mixing coefficient $\pi$ and parameter $\beta$ are optimized through the supervision of the depth as well as the color values, it improves the robustness of our MixNeRF for
slight changes of geometry.
Also, considering that the successful depth estimation is crucial to the rendered images' quality in a NeRF model \cite{niemeyer2022regnerf, deng2022depth, roessle2022dense, kim2022infonerf}, our direct estimation of the scene's depth benefits a lot.

\vspace{-4mm}
\paragraph{Blending weight regeneration.}
In addition, we exploit the estimated depth to regenerate the blending weights along the samples and model the RGB color values by a mixture of distributions once again.
Since the estimated depth of each sample is trained to be nearly identical to the ground truth depth, but not exactly the same, it can play a role of pseudo geometry for adjacent points of the sample without any additional pre-generation process of extra training data, \eg depth inputs made by SfM or rays from unobserved viewpoints.
The new blending weight $\hat{w}_j$ of the $j$-th sample along a ray based on the estimated depth $\mu_{\sh{j}}^d$ are defined as follow:
\begin{align}
\hat{w}_{j} = \hat{T}_{j}(1-\exp(-\sigma_{j}\hat{\delta}_{j})), \ 
\hat{\delta}_{j} = \mu^d_{j}(t_{j+1} - t_{j}),
\label{eq:additional_weight_gen}
\end{align}
in which we replace $\norm{\mathbf{d}}_2$ in $\delta_{j}$ formulation with $\mu_{j}^d$.
Finally, we model the color values along a ray based on the new mixing coefficients $\hat{\pi}$ derived from $\hat{w}$ and the corresponding pdf is as follows:
\begin{equation}
\hat{p}(\mathbf{c}|\mathbf{r}_i) = \sum_{j=1}^{M} \hat{\pi}_{ij}\mathcal{F}(\mathbf{c}; \mu_{ij}^\mathbf{c}, \beta_{ij}).
\label{eq:additional_mixture_pdf}
\end{equation}
Since the estimated ray depths are likely to be close enough to those of the ground truths, we use the same GT color values of input rays for modeling the mixture distribution based on the newly generated $\hat{\pi}$.
It further improves the robustness for shift of colors and ray viewpoints by simply modeling a ray once again with regenerated blending weights, eliminating pre-generation and extra inference of unseen views without much computational overhead.

\subsection{Total Loss}
\label{subsec:total_loss}
Our MixNeRF is trained to maximize the likelihood of $\mathbf{c}^\text{GT}$ and $d^\text{GT}$ for a set of input rays $\mathcal{R}$ as well as to minimize the $\mathcal{L}_\text{MSE}$.
Therefore, the loss functions can be simply defined to minimize the negative log-likelihood (NLL) of the ground truths as follows:
\begin{equation}
\begin{array}{cl}
\mathcal{L}^{C}_\text{NLL}= - \sum_{\mathbf{r} \in \mathcal{R}} \log p(\mathbf{c}^\text{GT}|\mathbf{r}), \\
\mathcal{L}^{D}_\text{NLL}= - \sum_{\mathbf{r} \in \mathcal{R}} \log p(d^\text{GT}|\mathbf{r}), \\
\hat{\mathcal{L}}^{C}_\text{NLL}= - \sum_{\mathbf{r} \in \mathcal{R}} \log \hat{p}(\mathbf{c}^\text{GT}|\mathbf{r}),
\label{eq:additional_mixture_loss}
\end{array}
\end{equation}
each of which corresponds to the NLL form of \cref{eq:mixture_pdf}, \cref{eq:dir_mixture_pdf} and \cref{eq:additional_mixture_pdf}, respectively.
As a result, we define our total loss as:
\begin{equation}
\mathcal{L}_\text{total} = \mathcal{L}_\text{MSE} + \lambda_{C}\mathcal{L}_\text{NLL}^{C} + \lambda_{D}\mathcal{L}_\text{NLL}^{D} + \hat{\lambda}_{C}\hat{\mathcal{L}}_\text{NLL}^{C},
\label{eq:total_loss}
\end{equation}
where $\lambda_C$, $\lambda_D$ and $\hat{\lambda}_C$ are balancing terms for the losses.
More details about training and implementation are provided in the supplementary material.

\section{Experiments}
\subsection{Experimental Details}
\paragraph{Datasets and metrics.}
We evaluate MixNeRF on the multiple standard benchmarks: DTU~\cite{jensen2014large}, LLFF~\cite{mildenhall2019local} and Realistic Synthetic 360$^\circ$~\cite{mildenhall2021nerf}.
DTU consists of images containing objects located on a white table with a black background.
LLFF contains real forward-facing scenes and is usually used as an out-of-distribution test set for pre-training methods.
We also compare our MixNeRF against other regularization methods on the Realistic Synthetic 360$^\circ$, which provides 8 synthetic scenes each consisting of 400 images rendered from inward-facing cameras with various viewpoints.
We follow the overall experimental protocols of \cite{yu2021pixelnerf, mildenhall2021nerf, kim2022infonerf, jain2021putting} for these datasets.

For the evaluation metrics, we adopt the mean of PSNR, structural similarity index (SSIM)~\cite{wang2004image}, LPIPS perceptual metric~\cite{zhang2018unreasonable}, and the geometric average~\cite{barron2021mip}.
Kindly refer to the supp. mat. for more details about datasets and metrics.

\vspace{-4mm}
\paragraph{Baselines.}
We compare our method against  several representative pre-training and regularization approaches~\cite{yu2021pixelnerf, chibane2021stereo, chen2021mvsnerf, jain2021putting, niemeyer2022regnerf, kim2022infonerf} as well as the vanilla mip-NeRF~\cite{barron2021mip}.
We report the evaluation results from \cite{niemeyer2022regnerf} for DTU and LLFF, which are superior to those from their original papers due to the improved training curriculum.
The DTU dataset is used as a pre-training resources for PixelNeRF~\cite{yu2021pixelnerf}, MVSNeRF~\cite{chen2021mvsnerf}, and SRF~\cite{chibane2021stereo}, and the LLFF dataset serves as an out-of-domain test set.
The regularization approaches and mip-NeRF are trained for each scene without pre-training.
For Realistic Synthetic 360$^\circ$, we train other regularization approaches~\cite{kim2022infonerf, jain2021putting, niemeyer2022regnerf} by their training schemes.
Note that the pre-trained RealNVP~\cite{dinh2016density} for training RegNeRF~\cite{niemeyer2022regnerf} is not publicly available and we report the results of RegNeRF trained without it on the Realistic Synthetic 360$^\circ$.
For the analysis of MixNeRF (\cref{sec:analysis}), all models including ours are trained with the same batch size and iterations.

\begin{figure}
    \centering
    \vspace{-.15cm}
    \begin{tabular}{p{0.19\textwidth}p{0.19\textwidth}}
     \centering\scriptsize RegNeRF~\cite{niemeyer2022regnerf} & \centering\scriptsize MixNeRF (Ours)
    \end{tabular}
     \begin{subfigure}[b]{0.45\textwidth}
         \centering
         \vspace{-.07cm}
        \includegraphics[width=\linewidth]{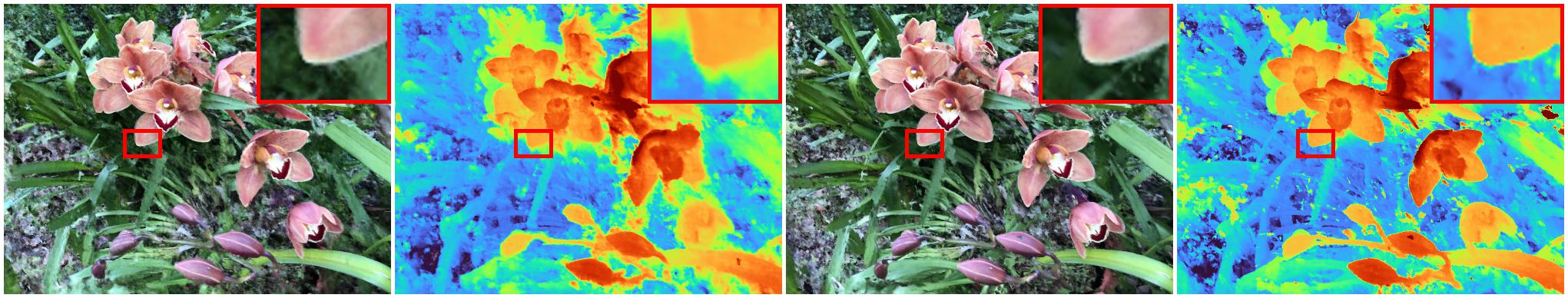}
        \vspace{-.52cm}
         \caption{\scriptsize LLFF 3-view}
     \end{subfigure}
     \begin{subfigure}[b]{0.45\textwidth}
         \centering
         \vspace{-.03cm}
        \includegraphics[width=\linewidth]{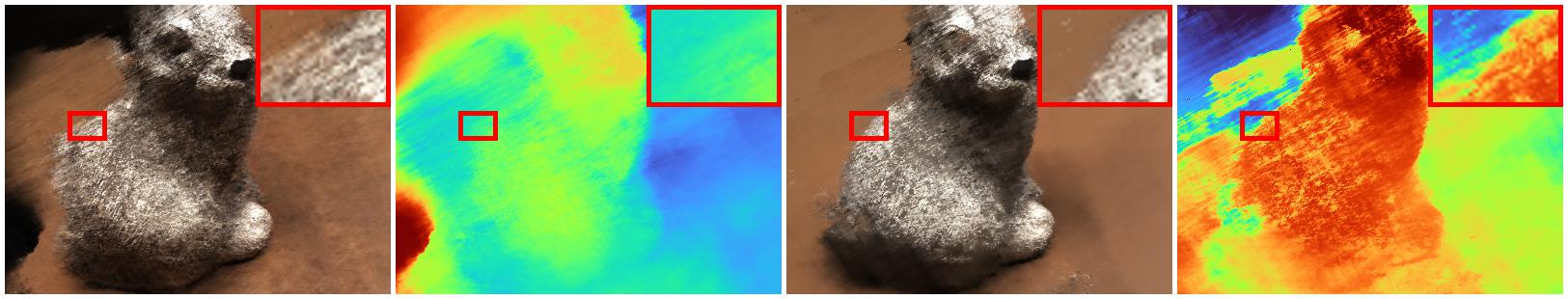}
        \vspace{-.52cm}
         \caption{\scriptsize DTU 3-view}
     \end{subfigure}
     \begin{subfigure}[b]{0.45\textwidth}
         \centering
         \vspace{-.03cm}
        \includegraphics[width=\linewidth]{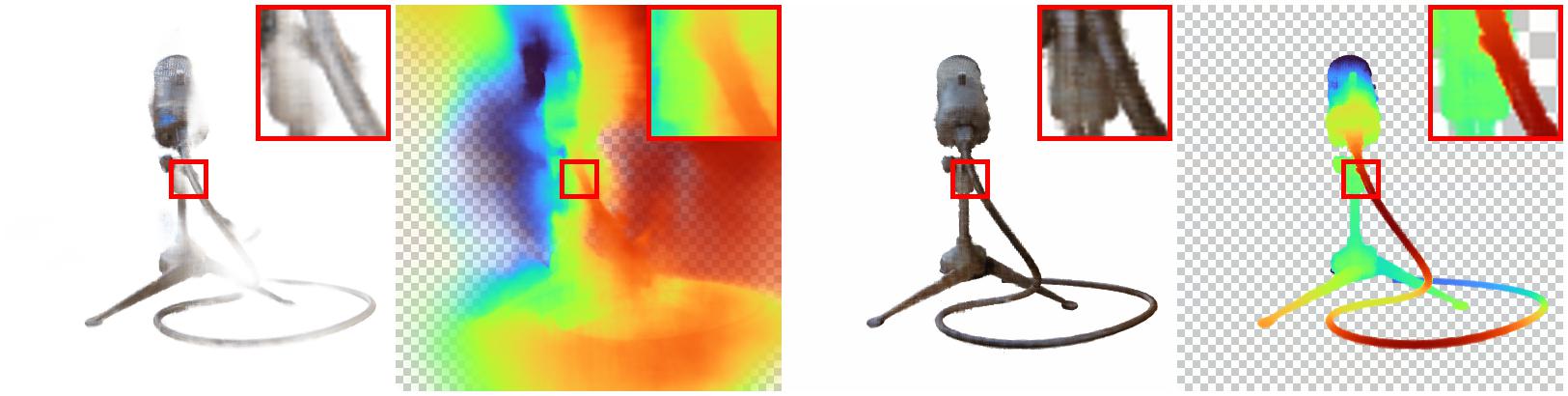}
        \vspace{-.52cm}
         \caption{\scriptsize Realistic Synthetic 360$^\circ$ 4-view}
     \end{subfigure}
     \vspace{-.2cm}
    \caption{
    \textbf{Comparison of estimated depth map.}
    We compare MixNeRF against RegNeRF, the state-of-the-art regularization approach.
    Our MixNeRF estimates more accurate depth maps and captures fine details better, leading to high-quality rendering with more distinct edges and less artifacts.
    }
    \label{fig:depth_vis}
    \vspace{-2mm}
\end{figure}

\subsection{Analysis of MixNeRF}
\label{sec:analysis}
\paragraph{Benefit of mixture density model.}
We leverage a mixture density model, which represents a complex multimodal distribution with a weighted combination of component distributions, to learn the distribution of density and colors along the ray samples effectively.
\cref{fig:weight_exp} demonstrates the comparison of the blending weight distributions of cast rays on the LLFF fern scene of 3-view scenario.
We compare ours against mip-NeRF and RegNeRF, with mip-NeRF trained from all training views as an ideal distribution.
For the unimodal distribution in blue, mip-NeRF does not estimate the mode well and achieves degenerate geometry.
However, RegNeRF and our MixNeRF show the unimodal weight distributions leading to  higher-quality novel views, and especially our MixNeRF achieves the distribution with sharper mode than RegNeRF, which is more similar to that of mip-NeRF (All-view).
In case of the bimodal-shaped distribution in red, our MixNeRF estimates the weight distribution successfully while both mip-NeRF and RegNeRF fail to estimate the accurate modes.
Since the predicted 3D geometry is directly correlated with how well the density is estimated, our MixNeRF is able to learn the geometry more efficiently with limited input views through mixture density modeling.

\vspace{1mm}
\noindent \textbf{Depth map estimation. }
We compare our MixNeRF against RegNeRF, which utilizes the prior of depth smoothness to learn 3D geometry, on the multiple benchmarks.
As illustrated in \cref{fig:depth_vis}, our MixNeRF estimates more accurate depth maps with distinct edges while RegNeRF generates both RGB images and depth maps with blurry fine details.
Especially for Realistic Synthetic 360$^\circ$, we observe that RegNeRF fails to learn the geometry with its smoothing strategy and achieves degenerate results due to the overly strong prior of depth smoothness.
However, since our MixNeRF learns the depth of a ray by leveraging a mixture density model without smoothing from additional unseen rays, the depth maps are predicted much more efficiently and precisely.

\begin{table}[!t]
\resizebox{\linewidth}{!}{
\begin{tabular}{l|c|c|c|c|c}
\toprule
\# of samples& 128 & 64 & 32 & 16 & 8 \\
\midrule
mip-NeRF~\cite{barron2021mip} & 0.332 & 0.331 & 0.329 & 0.308 & 0.251 \\
RegNeRF~\cite{niemeyer2022regnerf} & 0.587 & 0.585 & 0.576 & 0.522 & 0.379 \\
MixNeRF (Ours) & \textbf{0.629} & \textbf{0.629} & \textbf{0.620} & \textbf{0.580} & \textbf{0.468} \\
\bottomrule
\end{tabular}}
\vspace{-.2cm}
\caption{
    \textbf{Comparison with baselines by the number of ray samples.}
    Our MixNeRF with 75\% fewer samples (32-sample) outperforms RegNeRF with default 128-sample, and still achieves comparable results with only 16-sample ($\times$8 reduction).
    }
\vspace{-.2cm}
    \label{tab:inference_exp}
\end{table}

\begin{table*}[t]
\resizebox{\linewidth}{!}{
\begin{tabular}{l|c|ccc|ccc|ccc|ccc}
\toprule
  & \multirow{2}{*}{Approach} & \multicolumn{3}{c}{PSNR $\uparrow$} & \multicolumn{3}{c}{SSIM $\uparrow$} & \multicolumn{3}{c}{LPIPS $\downarrow$} & \multicolumn{3}{c}{Average Error $\downarrow$}  \\
  &  & 3-view & 6-view & 9-view  & 3-view & 6-view & 9-view  & 3-view & 6-view & 9-view  & 3-view & 6-view & 9-view \\
  \midrule
\multicolumn{14}{l}{\textit{\textbf{LLFF.}}} \\
\midrule\midrule
 mip-NeRF~\cite{barron2021mip} & N/A & 14.62 & 20.87 & 24.26 & 0.351 &0.692 & \cellcolor{yellow!25}0.805 & 0.495 & 0.255 & \cellcolor{yellow!25}0.172 & 0.246 & 0.114 & \cellcolor{yellow!25}0.073 \\
\midrule
PixelNeRF~\cite{yu2021pixelnerf} & \multirow{6}{*}{Pre-training} & 7.93 & 8.74 & 8.61 & 0.272 & 0.280 & 0.274 & 0.682 & 0.676 & 0.665 & 0.461 & 0.433 & 0.432 \\
PixelNeRF \textit{ft}~\cite{yu2021pixelnerf} & & 16.17 & 17.03 & 18.92 & 0.438 & 0.473 & 0.535 & 0.512 & 0.477 & 0.430 & 0.217 & 0.196 & 0.163 \\
SRF~\cite{chibane2021stereo} & & 12.34 & 13.10 & 13.00 & 0.250 & 0.293 & 0.297 & 0.591 & 0.594 & 0.605 & 0.313 & 0.293 & 0.296 \\
SRF \textit{ft}~\cite{chibane2021stereo} & & 17.07 & 16.75 & 17.39 & 0.436 & 0.438 & 0.465 & 0.529 & 0.521 & 0.503 & 0.203 & 0.207 & 0.193 \\
MVSNeRF~\cite{chen2021mvsnerf} &  & 17.25 & 19.79 & 20.47 & 0.557 & 0.656 & 0.689 & 0.356 & 0.269 & 0.242 & 0.171 & 0.125 & 0.111 \\
MVSNeRF \textit{ft}~\cite{chen2021mvsnerf} &  & \cellcolor{yellow!25}17.88 & 19.99 & 20.47 & \cellcolor{yellow!25}0.584 & 0.660 & 0.695 & \cellcolor{orange!25}0.327 & 0.264 & 0.244 & \cellcolor{yellow!25}0.157 & 0.122 & 0.111 \\
\midrule 
DietNeRF~\cite{jain2021putting} & \multirow{3}{*}{\shortstack{Regularization}} & 14.94 & \cellcolor{yellow!25}21.75 & \cellcolor{yellow!25}24.28 & 0.370 & \cellcolor{yellow!25}0.717 & 0.801 & 0.496 & \cellcolor{yellow!25}0.248 & 0.183 & 0.240 & \cellcolor{yellow!25}0.105 & \cellcolor{yellow!25}0.073 \\
RegNeRF~\cite{niemeyer2022regnerf} &  & \cellcolor{orange!25}19.08 & \cellcolor{orange!25}23.10 & \cellcolor{orange!25}24.86 & \cellcolor{orange!25}0.587 & \cellcolor{orange!25}0.760 & \cellcolor{orange!25}0.820 & \cellcolor{yellow!25}0.336 & \cellcolor{orange!25}0.206 & \cellcolor{orange!25}0.161 & \cellcolor{orange!25}0.146 & \cellcolor{orange!25}0.086 & \cellcolor{orange!25}0.067 \\
\textbf{MixNeRF (Ours)} &  & \cellcolor{red!25}\textbf{19.27} & \cellcolor{red!25}\textbf{23.76} & \cellcolor{red!25}\textbf{25.20} & \cellcolor{red!25}\textbf{0.629} & \cellcolor{red!25}\textbf{0.791} & \cellcolor{red!25}\textbf{0.833} & \cellcolor{red!25}\textbf{0.236} & \cellcolor{red!25}\textbf{0.115} & \cellcolor{red!25}\textbf{0.087} & \cellcolor{red!25}\textbf{0.124} & \cellcolor{red!25}\textbf{0.066} & \cellcolor{red!25}\textbf{0.052} \\
\midrule
\multicolumn{14}{l}{\textit{\textbf{DTU.}}} \\
\midrule\midrule
 mip-NeRF~\cite{barron2021mip} & N/A & 8.68 & 16.54 & 23.58 & 0.571 & 0.741 & \cellcolor{orange!25}0.879 & 0.353 & 0.198 & \cellcolor{yellow!25}0.092 & 0.323 & 0.148 & \cellcolor{yellow!25}0.056 \\
\midrule
PixelNeRF~\cite{yu2021pixelnerf} & \multirow{6}{*}{Pre-training} & 16.82 & 19.11 & 20.40 & 0.695 & 0.745 & 0.768 & 0.270 & 0.232 & 0.220 & 0.147 & 0.115 & 0.100 \\
PixelNeRF \textit{ft}~\cite{yu2021pixelnerf} & & \cellcolor{red!25}\textbf{18.95} & 20.56 & 21.83 & 0.710 & 0.753 & 0.781 & 0.269 & 0.223 & 0.203 & \cellcolor{yellow!25}0.125 & 0.104 & 0.090 \\
SRF~\cite{chibane2021stereo} & & 15.32 & 17.54 & 18.35 & 0.671 & 0.730 & 0.752 & 0.304 & 0.250 & 0.232 & 0.171 & 0.132 & 0.120 \\
SRF \textit{ft}~\cite{chibane2021stereo} & & 15.68 & 18.87 & 20.75 & 0.698 & 0.757 & 0.785 & 0.281 & 0.225 & 0.205 & 0.162 & 0.114 & 0.093 \\
MVSNeRF~\cite{chen2021mvsnerf} &  & \cellcolor{yellow!25}18.63 & \cellcolor{yellow!25}20.70 & 22.40 & \cellcolor{red!25}\textbf{0.769} & \cellcolor{yellow!25}0.823 & \cellcolor{yellow!25}0.853 & \cellcolor{orange!25}0.197 & 0.156 & 0.135 & \cellcolor{orange!25}0.113 & \cellcolor{yellow!25}0.088 & 0.068 \\
MVSNeRF \textit{ft}~\cite{chen2021mvsnerf} &  & 18.54 & 20.49 & 22.22 & \cellcolor{red!25}\textbf{0.769} & 0.822 & \cellcolor{yellow!25}0.853 & \cellcolor{orange!25}0.197 & \cellcolor{yellow!25}0.155 & 0.135 & \cellcolor{orange!25}0.113 & 0.089 & 0.069 \\
\midrule
DietNeRF~\cite{jain2021putting} & \multirow{3}{*}{\shortstack{Regularization}} & 11.85 & 20.63 & \cellcolor{yellow!25}23.83 & 0.633 & 0.778 & 0.823 & 0.314 & 0.201 & 0.173 & 0.243 & 0.101 & 0.068 \\
RegNeRF~\cite{niemeyer2022regnerf} &  & \cellcolor{orange!25}18.89 & \cellcolor{orange!25}22.20 & \cellcolor{orange!25}24.93 & \cellcolor{orange!25}0.745 & \cellcolor{red!25}\textbf{0.841} & \cellcolor{red!25}\textbf{0.884} & \cellcolor{red!25}\textbf{0.190} & \cellcolor{orange!25}0.117 & \cellcolor{orange!25}0.089 & \cellcolor{red!25}\textbf{0.112} & \cellcolor{orange!25}0.071 & \cellcolor{orange!25}0.047 \\
\textbf{MixNeRF (Ours)} &  & \cellcolor{red!25}\textbf{18.95} & \cellcolor{red!25}\textbf{22.30} & \cellcolor{red!25}\textbf{25.03} & \cellcolor{yellow!25}0.744 & \cellcolor{orange!25}0.835 & \cellcolor{orange!25}0.879 & \cellcolor{yellow!25}0.203 & \cellcolor{red!25}\textbf{0.102} & \cellcolor{red!25}\textbf{0.065} & \cellcolor{orange!25}0.113 & \cellcolor{red!25}\textbf{0.066} & \cellcolor{red!25}\textbf{0.042} \\
\bottomrule
\end{tabular}}
\vspace{-.2cm}
\caption{
    \textbf{Quantitative results on LLFF and DTU.}
    Our method achieves comparable or state-of-the-art performance across all scenarios in LLFF and DTU.
    For LLFF, our MixNeRF outperforms both pre-training and regularization baselines without any burdensome extra training resources.
    Likewise, MixNeRF achieves competitive results against the state-of-the-art methods on DTU dataset. 
    \textit{ft} indicates fine-tuning.
    }
    \label{tab:llff_dtu}
\vspace{-.3cm}
\end{table*}

\vspace{1mm}
\noindent \textbf{Efficiency in training and inference. } 
Our MixNeRF improves the efficiency for both the training and the inference phases by learning the 3D geometry  effectively without burdensome extra training resources.
\cref{fig:intro} illustrates that MixNeRF achieves superior performance with reduced training time among the vanilla mip-NeRF and two representative regularization methods on the LLFF.
For a fair comparison, we compare  the methods based on the identical JAX codebase~\cite{bradbury2018jax} using the same batch size and iterations on 2 NVIDIA TITAN RTX.
Although it takes a similar amount of time to train DietNeRF as MixNeRF, its performance is inferior significantly to ours in 3 and 6-view scenario.
Compared to RegNeRF, ours outperforms it with about 42\% shorter training time per scene under the same number of input view scenario, resulting from the elimination of extra inference for additional unseen rays.
Furthermore, we also observe that our MixNeRF shows better data efficiency requiring up to about 60\% fewer inputs than mip-NeRF to achieve comparable results, and outperforms mip-NeRF consistently in more than 9-view scenarios.
It indicates that our proposed training strategy is effective in general scenarios as well as the sparse input setting.
The related experimental results are provided in the suppl. material.
For the inference efficiency, \cref{tab:inference_exp} demonstrates the SSIM results by the number of samples along a ray on the LLFF under the 3-view scenario.
Our MixNeRF with 32-sample outperforms RegNeRF with default 128-sample, and still achieves comparable results with only 16-sample thanks to the capacity of our mixture model for representing the blending weight distributions successfully.

\begin{figure}[t]
\resizebox{\linewidth}{!}{
\begin{tabular}{c|ccc|c|c|c|c}
\toprule
& $\mathcal{L}_\text{NLL}^{C}$ & $\mathcal{L}_\text{NLL}^{D}$ & $\hat{\mathcal{L}}_\text{NLL}^{C}$ & {PSNR $\uparrow$} & {SSIM $\uparrow$} & {LPIPS $\downarrow$} & {Average $\downarrow$} \\
\midrule
(1) & \multicolumn{3}{c|}{Baseline (mip-NeRF${^\dagger}$~\cite{barron2021mip})} & 13.72 & 0.665 & 0.273 & 0.209 \\
\midrule\midrule
(2) & \checkmark & & & 17.44 & \cellcolor{yellow!25}0.729 & \cellcolor{red!25}\textbf{0.187} & \cellcolor{yellow!25}0.130 \\
(3) & \checkmark & \checkmark & & \cellcolor{orange!25}18.35 & \cellcolor{orange!25}0.739 & 0.209 & \cellcolor{orange!25}0.121 \\
(4) & \checkmark & & \checkmark & \cellcolor{yellow!25}17.45 & 0.728 & \cellcolor{orange!25}0.191 & 0.134 \\
(5) & \checkmark & \checkmark & \checkmark & \cellcolor{red!25}\textbf{18.95} & \cellcolor{red!25}\textbf{0.744} & \cellcolor{yellow!25}0.203 & \cellcolor{red!25}\textbf{0.113} \\
\bottomrule\bottomrule
\end{tabular}}
\vspace{.2cm}
\begin{subfigure}[b]{\linewidth}
\centering
\includegraphics[width=0.99\linewidth]{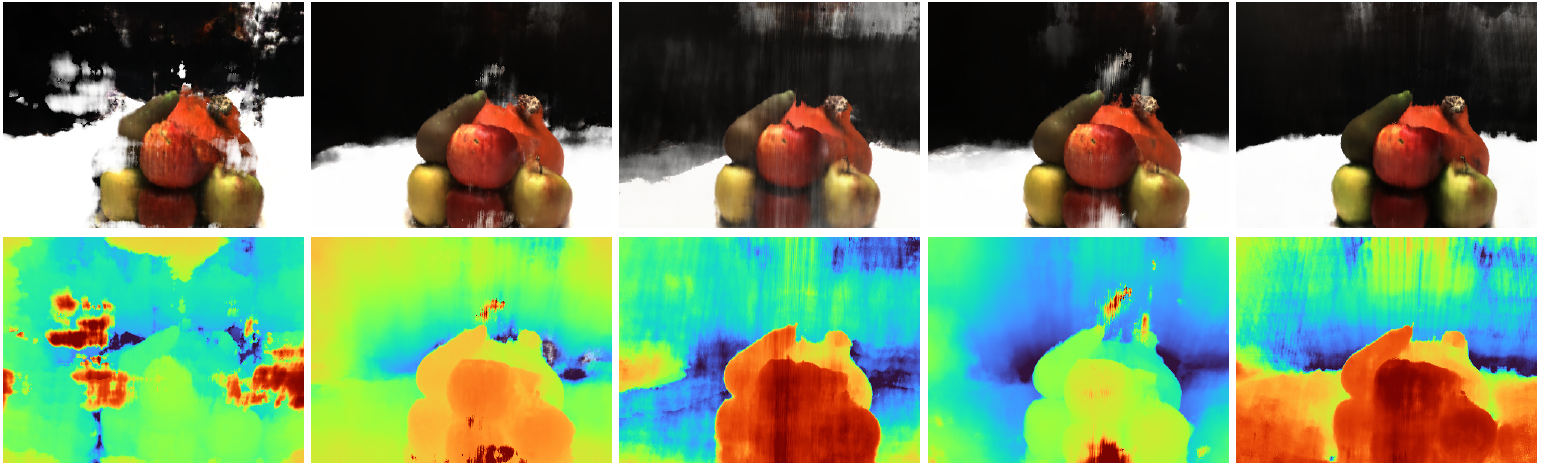}
\centering
\begin{tabular}{p{0.15\textwidth}p{0.15\textwidth}p{0.15\textwidth}p{0.15\textwidth}p{0.15\textwidth}}
     \vspace{-.4cm}\centering\scriptsize (1) & \vspace{-.4cm}\centering\scriptsize (2) & \vspace{-.4cm}\centering\scriptsize (3) & \vspace{-.4cm}\centering\scriptsize (4) & \vspace{-.4cm}\centering\scriptsize (5)
    \end{tabular}
\end{subfigure}
\vspace{-.8cm}
\caption{
\textbf{Ablation study.}
$\dagger$
uses a scene annealing strategy.
}
\vspace{-.2cm}
\label{fig:ablation}
\end{figure}

\subsection{Ablation Study}

We report the quantitative and qualitative results of our ablation study in \cref{fig:ablation}.
We observe that modeling a ray with mixture of distributions is helpful for improving performance under the sparse view setting ((1) $\rightarrow$ (2)).
Also, our proposed ray depth estimation task contributes to further improving the rendering quality by generating more accurate depth maps ((2) $\rightarrow$ (3)).
However, despite the well-estimated depth map, the RGB image suffers from the foggy artifacts upon the objects as shown in
(3).
By remodeling a ray through the weight regeneration process, our MixNeRF achieves high-quality of both RGB image and depth map ((3) $\rightarrow$ (5)).
Since the regenerated weights are not helpful without a supervision toward the ray depth estimation task ((2) $\rightarrow$ (4)), our proposed auxiliary task of ray depth estimation is useful for learning 3D geometry and playing a role of additional training resources by weight regeneration process on-the-fly.

\subsection{Comparison with other SOTA methods}

\begin{table}
\resizebox{\linewidth}{!}{
\begin{tabular}{l|cc|cc|cc|cc}
\toprule
  \multirow{2}{*}{} & \multicolumn{2}{c}{PSNR $\uparrow$} & \multicolumn{2}{c}{SSIM $\uparrow$} & \multicolumn{2}{c}{LPIPS $\downarrow$} & \multicolumn{2}{c}{Average Err. $\downarrow$}  \\
  & 4-view & 8-view & 4-view & 8-view & 4-view & 8-view & 4-view & 8-view \\ \midrule
mip-NeRF~\cite{barron2021mip} & 14.12 & 18.74 & 0.720 & 0.830 & 0.380 & 0.240 & 0.220 & 0.120 \\
\midrule
DietNeRF~\cite{jain2021putting} & \cellcolor{yellow!25}15.42 & \cellcolor{yellow!25}21.31 & 0.730 & \cellcolor{yellow!25}0.847 & \cellcolor{yellow!25}0.314 & \cellcolor{yellow!25}0.153 & \cellcolor{yellow!25}0.201 & \cellcolor{yellow!25}0.086 \\
InfoNeRF~\cite{kim2022infonerf} & \cellcolor{orange!25}18.44 & \cellcolor{orange!25}22.01 & \cellcolor{orange!25}0.792 & \cellcolor{orange!25}0.852 & \cellcolor{orange!25}0.223 & \cellcolor{orange!25}0.133 & \cellcolor{orange!25}0.119 & \cellcolor{orange!25}0.073 \\
RegNeRF~\cite{niemeyer2022regnerf} & 13.71 & 19.11 & \cellcolor{yellow!25}0.786 & 0.841 & 0.346 & 0.200 & 0.210 & 0.122 \\
\textbf{MixNeRF (Ours)} & \cellcolor{red!25}\textbf{18.99} & \cellcolor{red!25}\textbf{23.84} & \cellcolor{red!25}\textbf{0.807} & \cellcolor{red!25}\textbf{0.878} & \cellcolor{red!25}\textbf{0.199} & \cellcolor{red!25}\textbf{0.103} & \cellcolor{red!25}\textbf{0.113} & \cellcolor{red!25}\textbf{0.060} \\
\bottomrule
\end{tabular}}
\vspace{-.2cm}
\caption{
    \textbf{Quantitative results on Realistic Synthetic 360$^\circ$.}
    MixNeRF outperforms other state-of-the-art regularization methods by a large margin.
    }
    \label{tab:blender}
\vspace{-5mm}
\end{table}

\begin{figure*}
    \centering
    \begin{tabular}{p{0.07\textwidth}p{0.14\textwidth}p{0.14\textwidth}p{0.135\textwidth}p{0.14\textwidth}p{0.14\textwidth}p{0.07\textwidth}}
     & \centering\scriptsize mip-NeRF~\cite{barron2021mip} & \centering\scriptsize  DietNeRF~\cite{jain2021putting} & \centering\scriptsize  RegNeRF~\cite{niemeyer2022regnerf} & \centering\scriptsize  MixNeRF (Ours) & \centering\scriptsize Ground Truth &
    \end{tabular}
     \begin{subfigure}[b]{0.82\textwidth}
         \centering
        \vspace{-.07cm}
        \includegraphics[width=\linewidth]{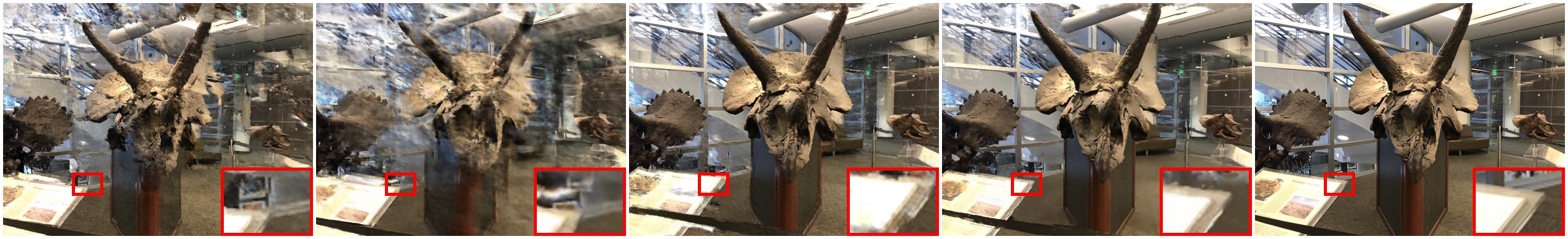}
        \vspace{-.52cm}
         \caption{\scriptsize LLFF 3-view}
         \label{fig:llff_3}
     \end{subfigure}
     \begin{subfigure}[b]{0.82\textwidth}
         \centering
         \vspace{-.03cm}
        \includegraphics[width=\linewidth]{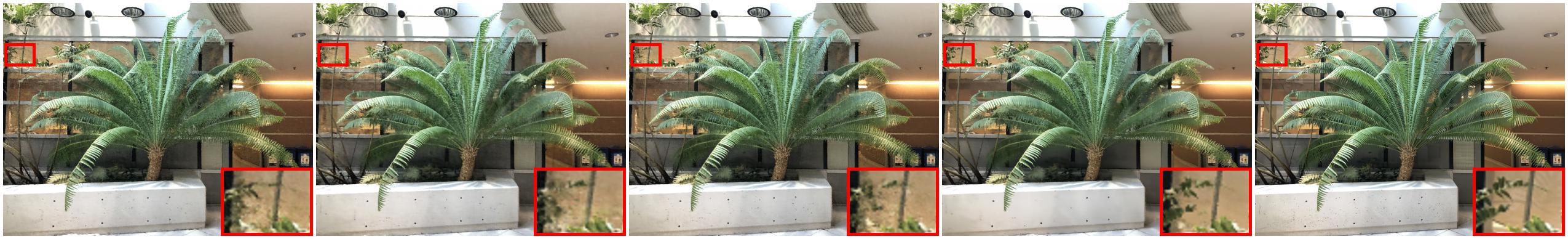}
        \vspace{-.52cm}
         \caption{\scriptsize LLFF 6-view}
         \label{fig:llff_6}
     \end{subfigure}
     \centering
    \begin{tabular}{p{0.07\textwidth}p{0.14\textwidth}p{0.14\textwidth}p{0.135\textwidth}p{0.14\textwidth}p{0.14\textwidth}p{0.07\textwidth}}
     & \centering\scriptsize mip-NeRF~\cite{barron2021mip} & \centering\scriptsize  PixelNeRF~\cite{yu2021pixelnerf} & \centering\scriptsize  RegNeRF~\cite{niemeyer2022regnerf} & \centering\scriptsize  MixNeRF (Ours) & \centering\scriptsize Ground Truth &
    \end{tabular}
     \begin{subfigure}[b]{0.82\textwidth}
         \centering
         \vspace{-.07cm}
        \includegraphics[width=\linewidth]{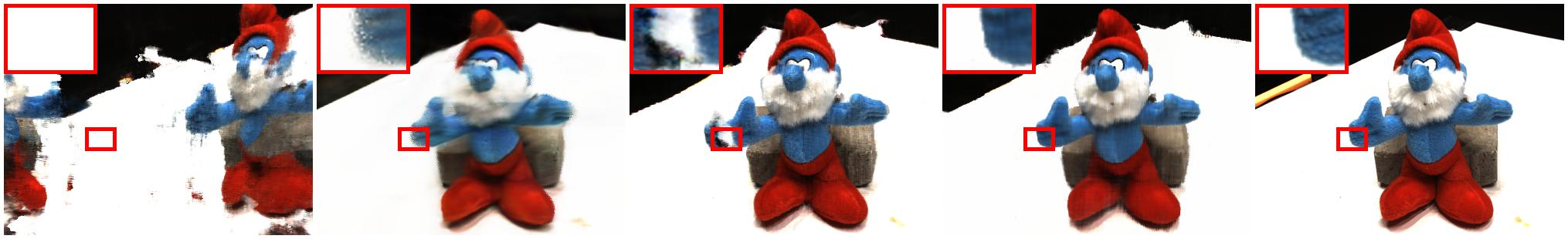}
        \vspace{-.52cm}
         \caption{\scriptsize DTU 3-view}
         \label{fig:dtu_3}
     \end{subfigure}
     \begin{subfigure}[b]{0.82\textwidth}
         \centering
         \vspace{-.03cm}
        \includegraphics[width=\linewidth]{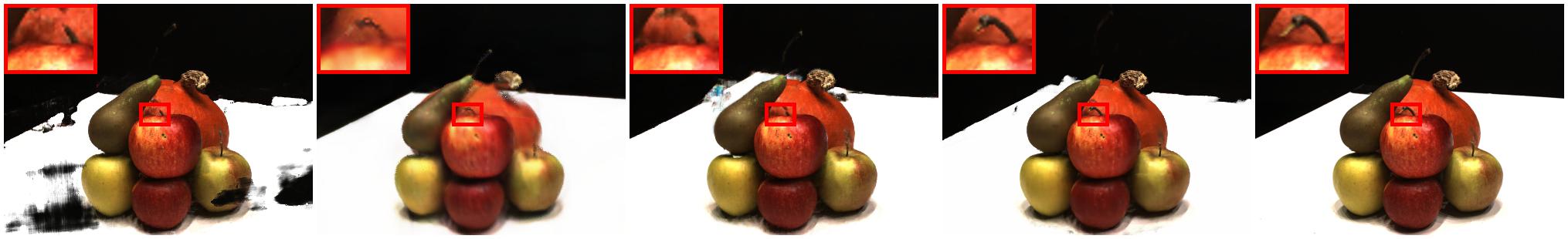}
        \vspace{-.52cm}
         \caption{\scriptsize DTU 6-view}
         \label{fig:dtu_6}
     \end{subfigure}
     \begin{tabular}{p{0.05\textwidth}p{0.12\textwidth}p{0.12\textwidth}p{0.11\textwidth}p{0.12\textwidth}p{0.12\textwidth}p{0.1\textwidth}p{0.22\textwidth}}
     & \centering\scriptsize mip-NeRF~\cite{barron2021mip} & \centering\scriptsize DietNeRF~\cite{jain2021putting} & \centering\scriptsize  InfoNeRF~\cite{kim2022infonerf} & \centering\scriptsize RegNeRF~\cite{niemeyer2022regnerf} & \centering\scriptsize MixNeRF (Ours) & \centering\scriptsize Ground Truth &
    \end{tabular}
     \begin{subfigure}[b]{0.85\textwidth}
         \centering
         \vspace{-.07cm}
        \includegraphics[width=\linewidth]{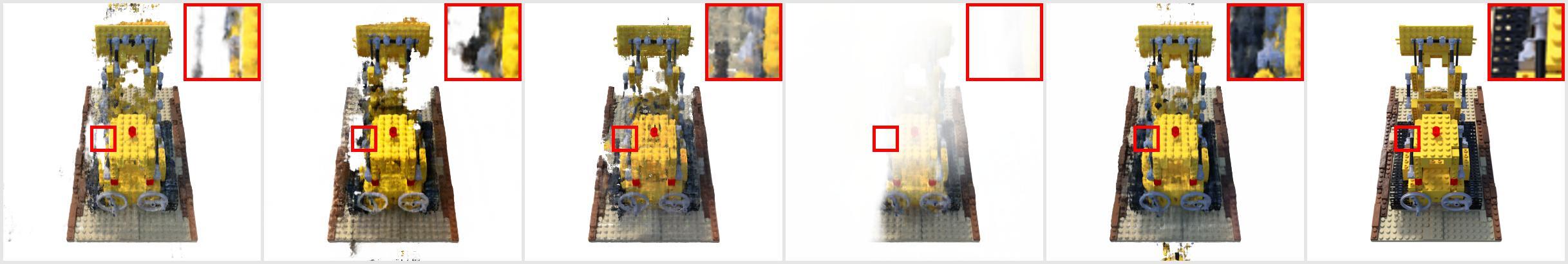}
        \vspace{-.52cm}
         \caption{\scriptsize Realistic Synthetic 360$^\circ$ 4-view}
         \label{fig:blender_4}
     \end{subfigure}
     \vspace{-.25cm}
    \caption{
    \textbf{Qualitative results on LLFF (a,b), DTU (c,d), and Realistic Synthetic 360$^\circ$ (e).}
    More results are provided in suppl. material.
    }
    \label{fig:qual_res}
     \vspace{-.4cm}
\end{figure*}

\paragraph{LLFF.}
As shown in \cref{tab:llff_dtu}, the pre-training approaches except MVSNeRF are not able to achieve comparable results without fine-tuning for the 3-view scenario.
The regularization approaches and vanilla mip-NeRF outperform the pre-training approaches in 6 and 9-view settings.
Especially, RegNeRF improves the rendering quality by a large margin compared to mip-NeRF and DietNeRF, thanks to the regularization strategy using pre-generated rays from unseen viewpoints.
Our MixNeRF achieves state-of-the-art results across all scenarios and metrics without any pre-training process using a large-scale dataset or extra inference for pre-generated training resources.
It improves the training efficiency, \eg requiring about 42\% shorter training time than RegNeRF in the same input view setting (see \cref{fig:intro}).
Furthermore, as demonstrated in \cref{fig:qual_res}, we observe that our method achieves more realistic results with fine details than RegNeRF since MixNeRF learns the 3D geometry successfully without an explicit regularization for smoothing the depth.

\noindent \textbf{DTU. }
Since the pre-training approaches are conditioned from the prior knowledge of DTU, they achieve competitive results with and without fine-tuning as demonstrated in \cref{tab:llff_dtu}.
Our MixNeRF achieves comparable or better results than other baselines.
Compared to the PSNR, our MixNeRF achieves worse quantitative results than RegNeRF for SSIM.
We conjecture that since RegNeRF uses an explicit smoothing term for regularizing the depth, it can achieve slightly better quantitative results of SSIM which is a patch-wise evaluation metric.
However, as shown in \cref{fig:qual_res}, our MixNeRF renders clearer images, capturing fine details better than other baselines despite a little worse quantitative results for some metrics.
More qualitative results of LLFF and DTU are provided in the suppl. material.

\noindent \textbf{Realistic Synthetic 360$^\circ$. }
As shown in \cref{tab:blender}, our MixNeRF outperforms other regularization baselines across all settings and metrics by a large margin.
\cref{fig:blender_4} illustrates that other methods suffer from  severe floating artifacts and degenerate colors in the 4-view scenario.
Especially, the depth smoothing strategy of RegNeRF rather brings about the significant performance degradation.
It implies that smoothing is not a fundamental solution universally effective for different datasets.
Compared to the baselines, our MixNeRF achieves superior rendering quality with much less artifacts and more accurate geometry in both 4 and 8-view (see the suppl. material) scenarios.

\section{Conclusion}
We have introduced MixNeRF, a novel regularization approach for training NeRF in the limited data scenario.
However, previous approaches heavily depend on the extra training resources, which goes against the philosophy of sparse-input novel-view synthesis pursuing the efficiency of training.  To overcome this bottleneck, we propose modeling a ray with mixture density, which enables effective learning of 3D geometry with sparse inputs.
Furthermore, our novel training strategy, consisting of an auxiliary ray depth estimation and the following weight regeneration, further improves the rendering quality and better reconstructs 3D geometry by more accurate depth estimation without any extra training resources that should be prepared in advance.
Our proposed MixNeRF outperforms both pre-training and regularization approaches across the multiple  benchmarks with an enhanced efficiency of training and inference.

{\small
\bibliographystyle{ieee_fullname}
\bibliography{egbib}
}

\newpage

\setcounter{section}{0}
\setcounter{table}{0}
\setcounter{figure}{0}
\renewcommand\thesection{\Alph{section}}
\renewcommand\thetable{\Alph{table}}
\renewcommand\thefigure{\Alph{figure}}

\twocolumn[{
\centering
\resizebox{0.8\linewidth}{!}{
\begin{tabular}{c|c|ccc|ccc|cc}
\toprule
  & \multirow{2}{*}{Anneal.} & \multicolumn{3}{c}{LLFF~\cite{mildenhall2019local}} & \multicolumn{3}{c}{DTU~\cite{jensen2014large}} & \multicolumn{2}{c}{Real. Syn. 360$^\circ$~\cite{mildenhall2021nerf}}  \\
  &  & 3-view & 6-view & 9-view & 3-view & 6-view & 9-view & 4-view & 8-view \\
  \midrule
 $\lambda_C$ & \checkmark & \multicolumn{8}{c}{$[4.0, 1\mathrm{e}{-3}]$} \\
 \midrule
 $\lambda_D$ & & $1\mathrm{e}{-4}$ & $1\mathrm{e}{-5}$ & $1\mathrm{e}{-6}$ & $1\mathrm{e}{-3}$ & $1\mathrm{e}{-4}$ & $1\mathrm{e}{-5}$ & $1\mathrm{e}{-3}$ & $1\mathrm{e}{-4}$ \\
 \midrule
 $\hat{\lambda}_C$ & & $1\mathrm{e}{-5}$ & $1\mathrm{e}{-6}$ & $1\mathrm{e}{-7}$ & $1\mathrm{e}{-4}$ & $1\mathrm{e}{-5}$ & $1\mathrm{e}{-6}$ & $1\mathrm{e}{-4}$ & $1\mathrm{e}{-5}$ \\
\bottomrule
\end{tabular}}
\captionof{table}{
    \textbf{Overview of our loss balancing weights.}
    We apply a linear annealing strategy for $\lambda_C$ to stabilize the training.
    We divide $\lambda_D$ and $\hat{\lambda}_C$ by a factor of 10 as more input views are provided for training.
    }
    \label{tab:loss_weights}
    \vspace{.5cm}
}]

\section{Implementation Details}
\subsection{Hyperparameters}

Following RegNeRF~\cite{niemeyer2022regnerf}, we adopt a scene space annealing during the early training stage, an exponential learning rate decay from $2\mathrm{e}{-3}$ to $2\mathrm{e}{-5}$, and 512 steps of warm up~\cite{barron2021mip} with a delay multiplier of $1\mathrm{e}{-2}$.
For the Realistic Synthetic 360$^\circ$~\cite{mildenhall2021nerf}, we set the initial learning rate as $1\mathrm{e}{-3}$ and apply an exponential decay to $1\mathrm{e}{-5}$.
The Adam~\cite{kingma2014adam} optimizer is used and the gradient clippings are applied by value at 0.1 and norm at 0.1 in order.
We train our MixNeRF for 500 pixel epochs with 4096 batch size on 2 NVIDIA TITAN RTX, and the training time is measured on the same hardware.
For the balancing hyperparameters for our loss terms, we anneal $\lambda_C$ from $4.0$ to $1\mathrm{e}{-3}$ over the first 512 iterations, while setting $\lambda_D$ and $\hat{\lambda}_C$ as different values by the datasets.
\cref{tab:loss_weights} shows the overview of balancing terms by the datasets and the number of input views.

\subsection{Architecture}

\begin{figure}[t]
\centering
\includegraphics[width=1.0\linewidth]{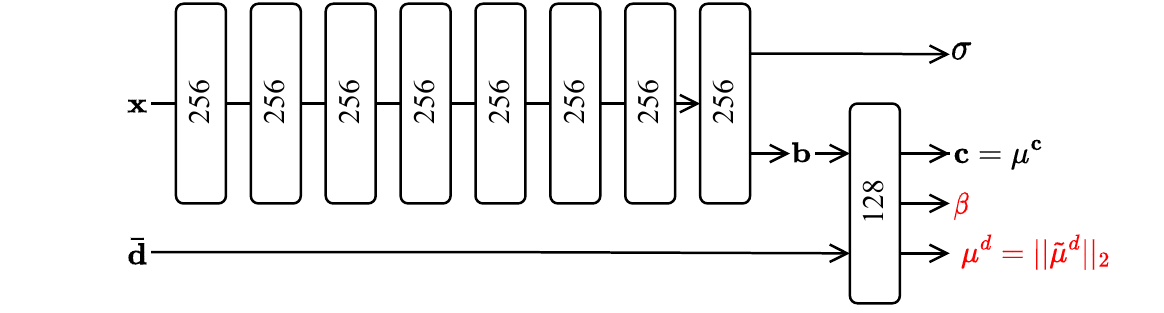}
\caption{\textbf{MixNeRF Network Architecture.} The architecture of MixNeRF is implemented based upon the mip-NeRF~\cite{barron2021mip}.
It additionally outputs the scale parameter $\beta$ using softplus activation and the ray depths $\mu^d = \norm{\tilde{\mu}^d}_2$, which are denoted in red.
$\mathbf{b}$ indicates a bottleneck vector.
}
\vspace{-.3cm}
\label{fig:supp_arch}
\end{figure}

Our MixNeRF is based on the architecture of mip-NeRF~\cite{barron2021mip}.
As illustrated in \cref{fig:supp_arch}, our MixNeRF additionally outputs the scale parameters $\beta$ using softplus activation and the ray depths $\mu^d$ for our mixture model.
In practice, we estimate the unnormalized ray directions $\tilde{\mu}^d \in \mathbb{R}^{\text{N} \times 3}$, where $N$ indicates the number of samples, and we use its Euclidean norm $\mu^d = \norm{\tilde{\mu}^d}_2$ as the estimated ray depths for the training stability.

\section{Experimental Details}
\subsection{Datasets}
We evaluate MixNeRF on the different standard benchmarks: LLFF~\cite{mildenhall2019local}, DTU~\cite{jensen2014large}, and Realistic Synthetic 360$^\circ$~\cite{mildenhall2021nerf}.
\paragraph{LLFF:}
It contains realistic forward-facing scenes and is generally used as an out-of-domain test set for pre-training methods.
Following the protocol of \cite{mildenhall2021nerf}, every 8-th image is used as a held-out test set and input views are chosen evenly from the remaining images.
We report the results under the scenarios of 3, 6, and 9 input views following \cite{yu2021pixelnerf}.

\paragraph{DTU:}
It consists of images containing objects placed on a white table with a black background.
We follow the experimental protocol of \cite{yu2021pixelnerf} and conduct experiments on their designated 15 scenes.
As with the LLFF dataset, we conduct the experiments under the scenarios of 3, 6, and 9-view.

\paragraph{Realistic Synthetic 360$^\circ$:}
It consists of 8 inward-facing synthetic scenes with different viewpoints, each containing 400 images.
Following previous works~\cite{kim2022infonerf, jain2021putting}, we conduct the experiments for the scenarios of 4 and 8 views.
For a fair comparison with other regularization methods, we sample the first 4 and 8 images from the training set for the scenario of 4 and 8 input views, respectively, for all models and use the 200 test set images for evaluation.
Note that the images of the training set are arranged randomly in the first place, and we do not choose the training input views carefully for improving the performance.

\subsection{Evaluation Metrics}
We adopt a set of evaluation metrics including the mean of PSNR, structural similarity index (SSIM)~\cite{wang2004image}, and LPIPS perceptual metric~\cite{zhang2018unreasonable}.
Additionally, we report its geometric average~\cite{barron2021mip}: $\text{MSE} = 10^{-\text{PSNR}/10}$, $\sqrt{1-\text{SSIM}}$, and LPIPS.
Following \cite{niemeyer2022regnerf}, we adopt masked metrics to avoid background bias for DTU.

\section{Data Efficiency Experiment}

\begin{figure}[t]
\centering
\includegraphics[width=1.0\linewidth]{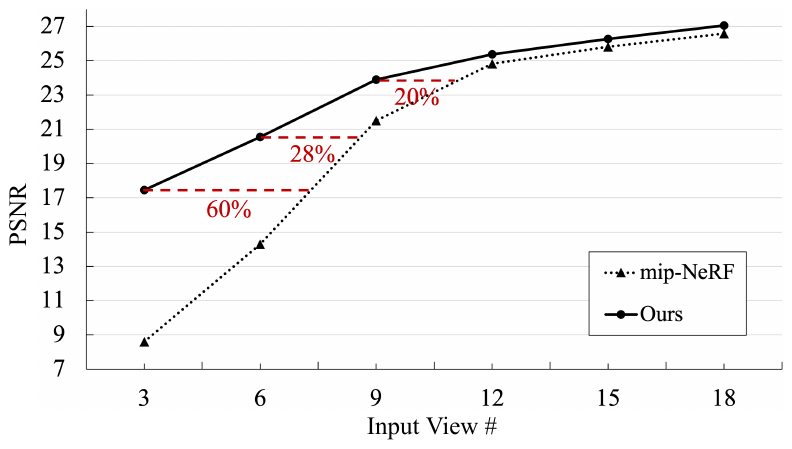}
\vspace{-.7cm}
\caption{
\textbf{Comparison with baseline by the number of input views.}
Our MixNeRF requires up to about 60\% fewer input views than mip-NeRF to achieve comparable performance, and outperforms mip-NeRF consistently even when more input views are used for training.
Since the reduced test set is used for the experiment following \cite{niemeyer2022regnerf}, the results can be slightly different from the main table.
}
\vspace{-.1cm}
\label{fig:data_efficiency}
\end{figure}

As demonstrated in \cref{fig:data_efficiency}, we observe that our MixNeRF achieves superior data efficiency to the vanilla mip-NeRF.
Our MixNeRF requires up to about 60\% fewer input views to mip-NeRF to achieve comparable results.
Moreover, ours outperforms mip-NeRF consistently even when more than 9 input views are provided.
It indicates that our proposed mixture modeling strategy is effective in general scenarios as well as the sparse input setting.

\section{Additional Qualitative Results}

\begin{figure*}
    \centering
    \begin{tabular}{p{0.07\textwidth}p{0.14\textwidth}p{0.14\textwidth}p{0.135\textwidth}p{0.14\textwidth}p{0.14\textwidth}p{0.07\textwidth}}
     & \centering\scriptsize mip-NeRF~\cite{barron2021mip} & \centering\scriptsize  DietNeRF~\cite{jain2021putting} & \centering\scriptsize  RegNeRF~\cite{niemeyer2022regnerf} & \centering\scriptsize  MixNeRF (Ours) & \centering\scriptsize Ground Truth &
    \end{tabular}
    \begin{subfigure}[b]{0.82\textwidth}
         \centering
        \vspace{-.07cm}
        \includegraphics[width=\linewidth]{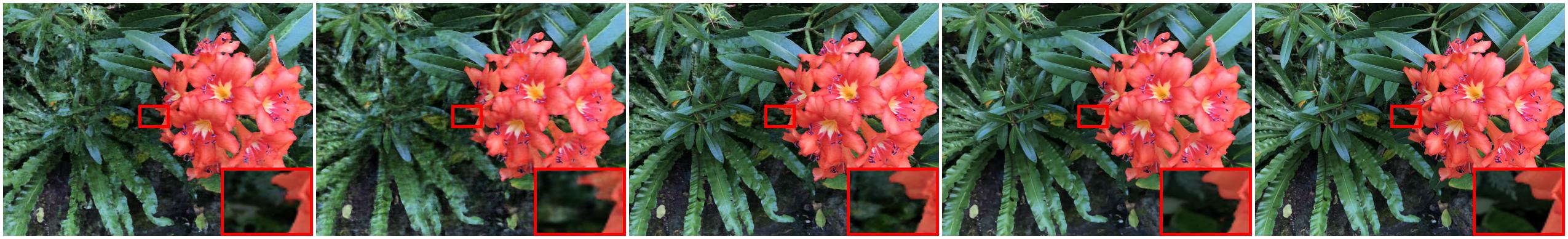}
        \vspace{-.52cm}
         \caption{\scriptsize 3-view}
         \label{fig:supp_llff_3}
     \end{subfigure}
     \begin{subfigure}[b]{0.82\textwidth}
         \centering
        \vspace{-.07cm}
        \includegraphics[width=\linewidth]{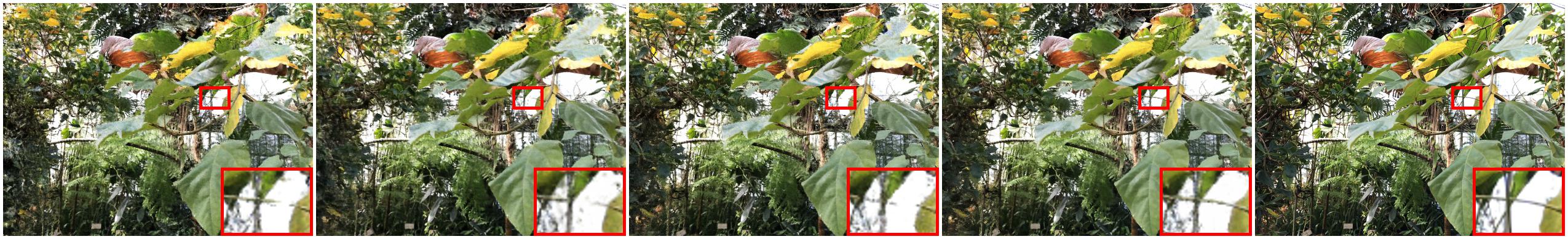}
        \vspace{-.52cm}
         \caption{\scriptsize 6-view}
         \label{fig:supp_llff_6}
     \end{subfigure}
     \begin{subfigure}[b]{0.82\textwidth}
         \centering
        \vspace{-.07cm}
        \includegraphics[width=\linewidth]{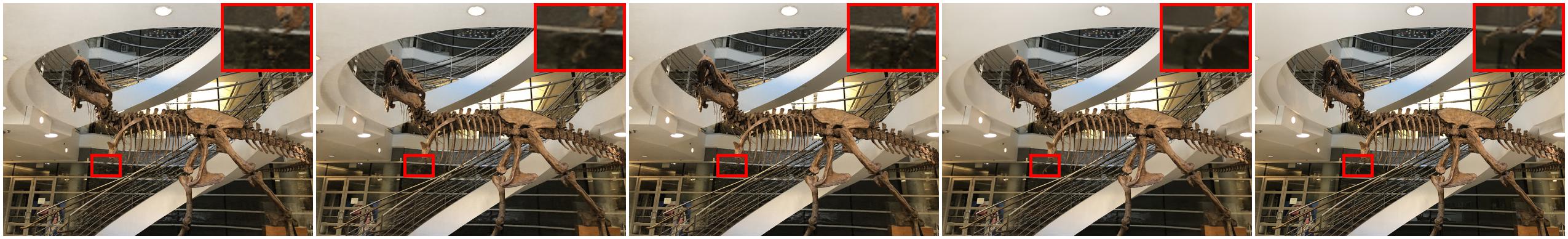}
        \vspace{-.52cm}
         \caption{\scriptsize 9-view}
         \label{fig:supp_llff_9}
     \end{subfigure}
     \vspace{-.3cm}
    \caption{
    \textbf{Additional qualitative comparisons on LLFF.}
    }
    \label{fig:supp_qual_res_llff}
     \vspace{-.3cm}
\end{figure*}
\begin{figure*}
     \centering
    \begin{tabular}{p{0.07\textwidth}p{0.14\textwidth}p{0.14\textwidth}p{0.135\textwidth}p{0.14\textwidth}p{0.14\textwidth}p{0.07\textwidth}}
     & \centering\scriptsize mip-NeRF~\cite{barron2021mip} & \centering\scriptsize  PixelNeRF~\cite{yu2021pixelnerf} & \centering\scriptsize  RegNeRF~\cite{niemeyer2022regnerf} & \centering\scriptsize  MixNeRF (Ours) & \centering\scriptsize Ground Truth &
    \end{tabular}
     \begin{subfigure}[b]{0.82\textwidth}
         \centering
         \vspace{-.07cm}
        \includegraphics[width=\linewidth]{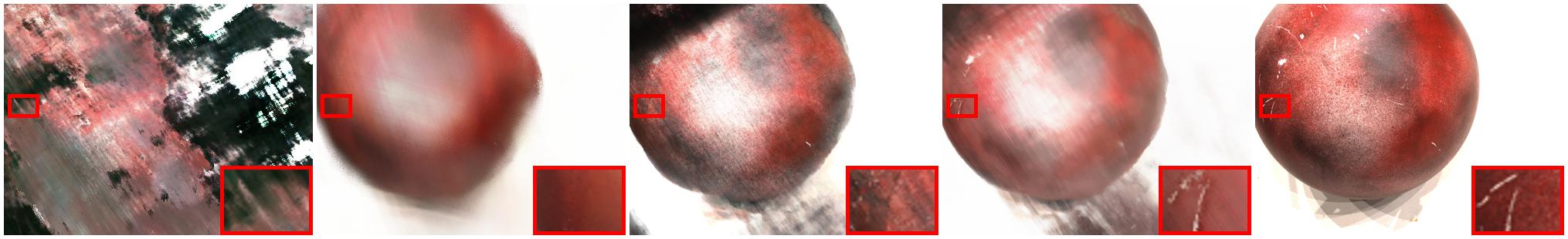}
        \vspace{-.52cm}
         \caption{\scriptsize 3-view}
         \label{fig:supp_dtu_3}
     \end{subfigure}
     \begin{subfigure}[b]{0.82\textwidth}
         \centering
         \vspace{-.07cm}
        \includegraphics[width=\linewidth]{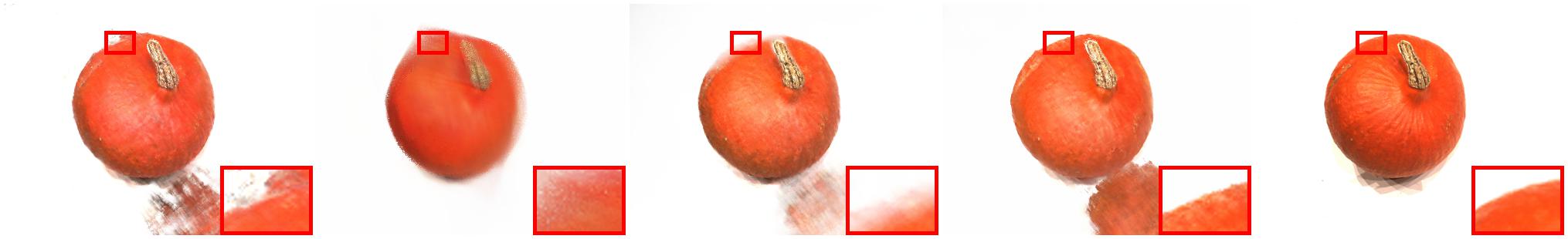}
        \vspace{-.52cm}
         \caption{\scriptsize 6-view}
         \label{fig:supp_dtu_6}
     \end{subfigure}
     \begin{subfigure}[b]{0.82\textwidth}
         \centering
         \vspace{-.07cm}
        \includegraphics[width=\linewidth]{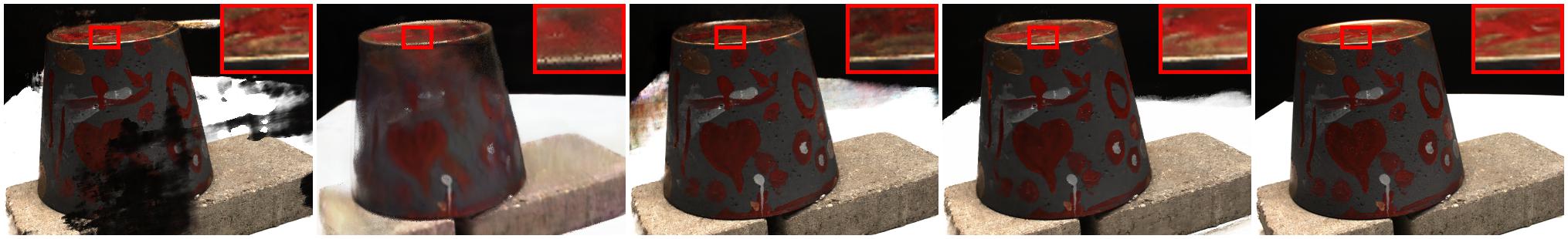}
        \vspace{-.52cm}
         \caption{\scriptsize 9-view}
         \label{fig:supp_dtu_9}
     \end{subfigure}
     \vspace{-.3cm}
    \caption{
    \textbf{Additional qualitative comparisons on DTU.}
    }
    \label{fig:supp_qual_res_dtu}
     \vspace{-.3cm}
\end{figure*}
\begin{figure*}
\centering
     \begin{tabular}{p{0.05\textwidth}p{0.12\textwidth}p{0.12\textwidth}p{0.11\textwidth}p{0.12\textwidth}p{0.12\textwidth}p{0.1\textwidth}p{0.22\textwidth}}
     & \centering\scriptsize mip-NeRF~\cite{barron2021mip} & \centering\scriptsize DietNeRF~\cite{jain2021putting} & \centering\scriptsize  InfoNeRF~\cite{kim2022infonerf} & \centering\scriptsize RegNeRF~\cite{niemeyer2022regnerf} & \centering\scriptsize MixNeRF (Ours) & \centering\scriptsize Ground Truth &
    \end{tabular}
    \begin{subfigure}[b]{0.83\textwidth}
         \centering
         \vspace{-.07cm}
        \includegraphics[width=\linewidth]{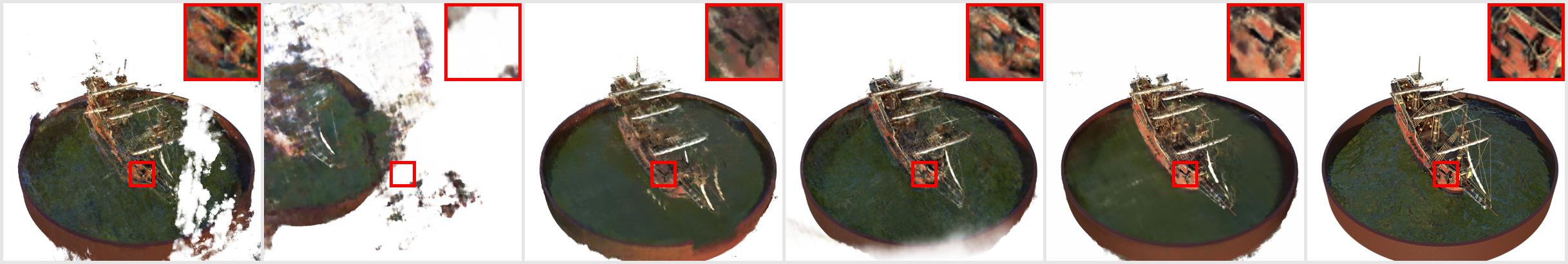}
        \vspace{-.52cm}
         \caption{\scriptsize 4-view}
         \label{fig:supp_blender_4}
     \end{subfigure}
     \begin{subfigure}[b]{0.83\textwidth}
         \centering
        \includegraphics[width=\linewidth]{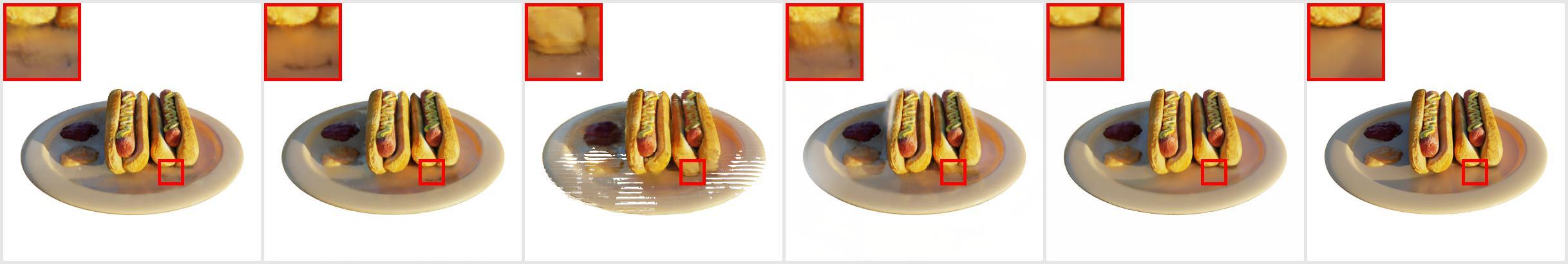}
        \vspace{-.52cm}
         \caption{\scriptsize 8-view}
         \label{fig:supp_blender_8}
     \end{subfigure}
     \vspace{-.3cm}
    \caption{
    \textbf{Additional qualitative comparisons on Realistic Synthetic 360$^\circ$.}
    }
    \label{fig:supp_qual_res_blender}
     \vspace{-.4cm}
\end{figure*}

\begin{figure*}
     \centering
     \begin{subfigure}[b]{0.8\textwidth}
         \centering
         \vspace{-.07cm}
        \includegraphics[width=\linewidth]{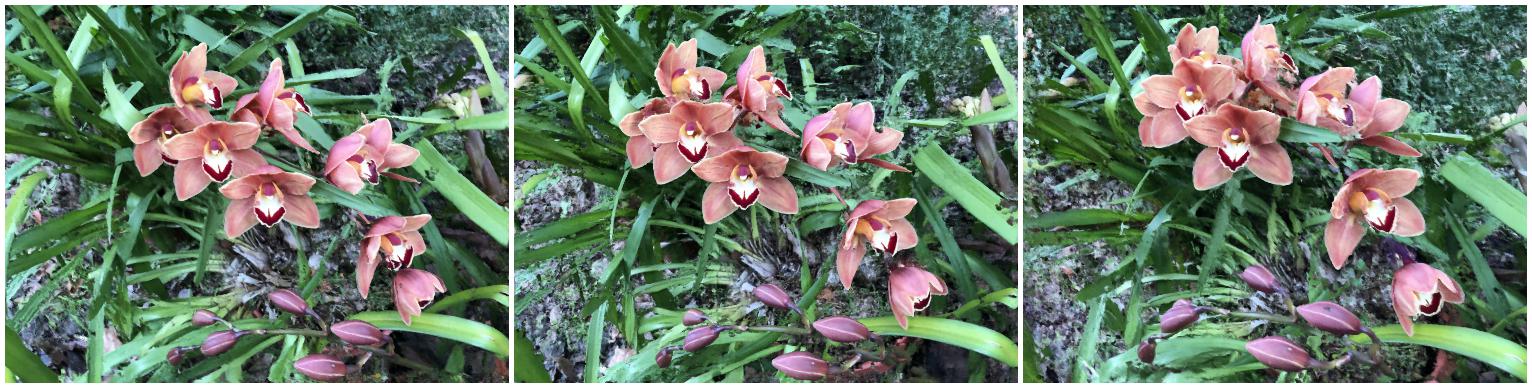}
        \vspace{-.52cm}
         \label{fig:supp_llff_3_ours_1}
     \end{subfigure}
     \begin{subfigure}[b]{0.8\textwidth}
         \centering
         \vspace{-.07cm}
        \includegraphics[width=\linewidth]{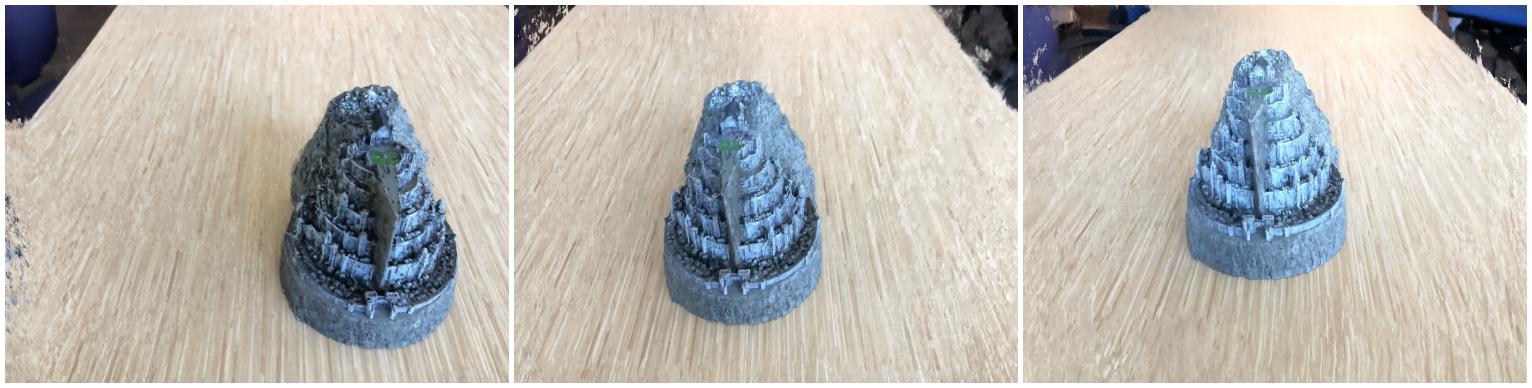}
        \vspace{-.52cm}
        \caption{\scriptsize 3-view}
         \label{fig:supp_llff_3_ours_2}
     \end{subfigure}
     \begin{subfigure}[b]{0.8\textwidth}
         \centering
         \vspace{-.07cm}
        \includegraphics[width=\linewidth]{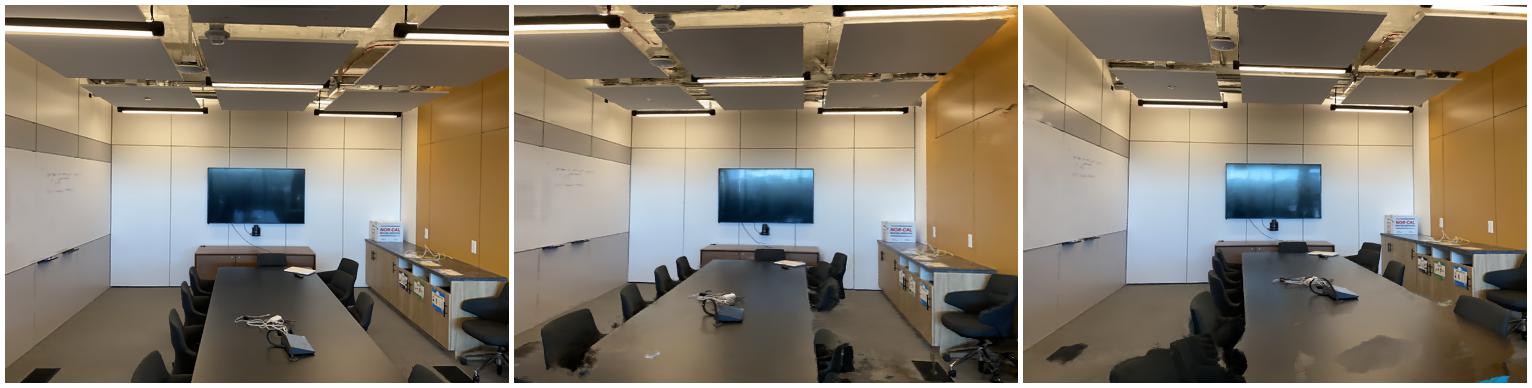}
        \vspace{-.52cm}
         \label{fig:supp_llff_6_ours_1}
     \end{subfigure}
     \begin{subfigure}[b]{0.8\textwidth}
         \centering
         \vspace{-.07cm}
        \includegraphics[width=\linewidth]{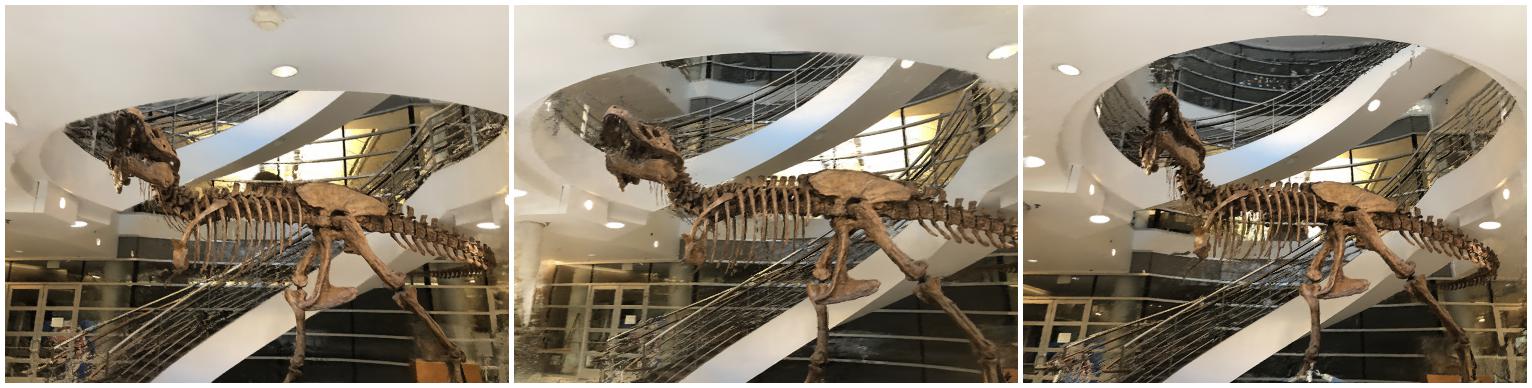}
        \vspace{-.52cm}
        \caption{\scriptsize 6-view}
         \label{fig:supp_llff_6_ours_2}
     \end{subfigure}
     \begin{subfigure}[b]{0.8\textwidth}
         \centering
         \vspace{-.07cm}
        \includegraphics[width=\linewidth]{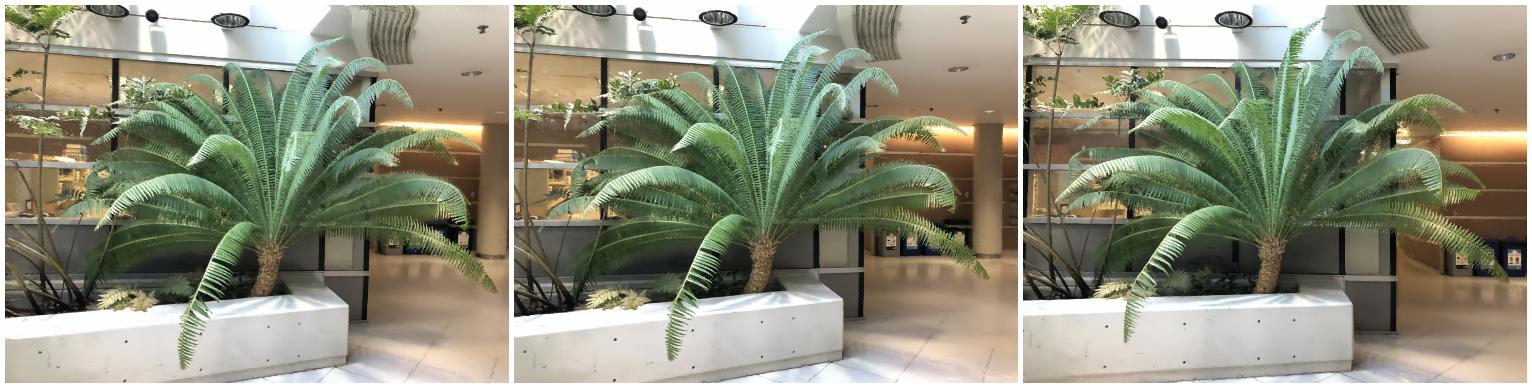}
        \vspace{-.52cm}
         \label{fig:supp_llff_9_ours_1}
     \end{subfigure}
     \begin{subfigure}[b]{0.8\textwidth}
         \centering
         \vspace{-.07cm}
        \includegraphics[width=\linewidth]{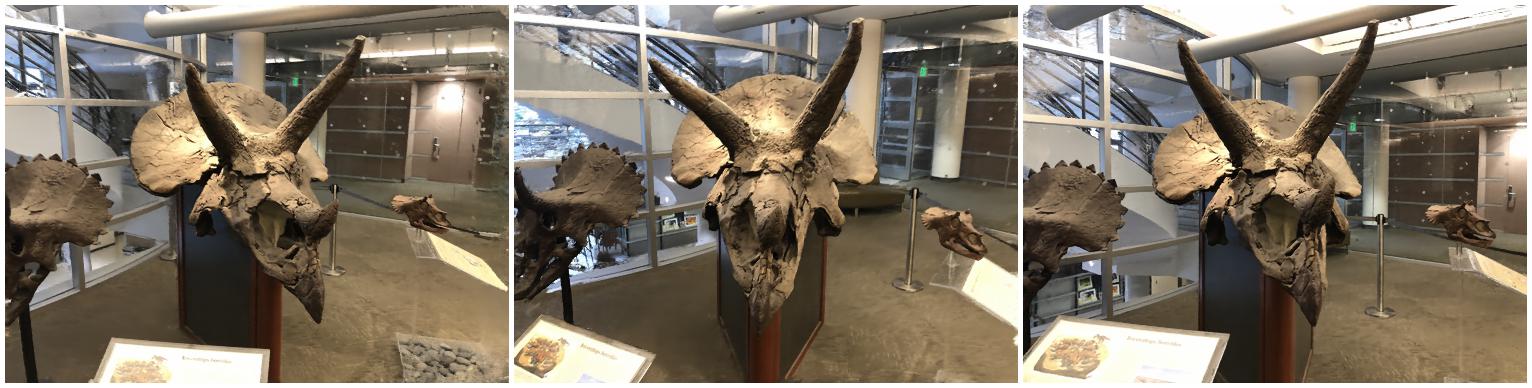}
        \vspace{-.52cm}
        \caption{\scriptsize 9-view}
         \label{fig:supp_llff_9_ours_2}
     \end{subfigure}
     \vspace{-.25cm}
    \caption{
    \textbf{Additional qualitative results of our MixNeRF on LLFF.}
    }
    \label{fig:supp_qual_res_llff_ours}
\end{figure*}

\begin{figure*}
     \centering
     \begin{subfigure}[b]{0.9\textwidth}
         \centering
         \vspace{-.07cm}
        \includegraphics[width=\linewidth]{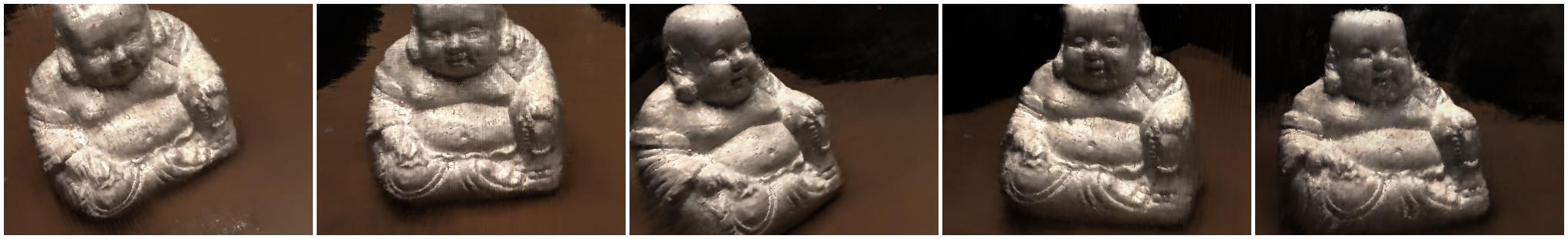}
        \vspace{-.52cm}
         \label{fig:supp_dtu_3_ours_1}
     \end{subfigure}
     \begin{subfigure}[b]{0.9\textwidth}
         \centering
         \vspace{-.07cm}
        \includegraphics[width=\linewidth]{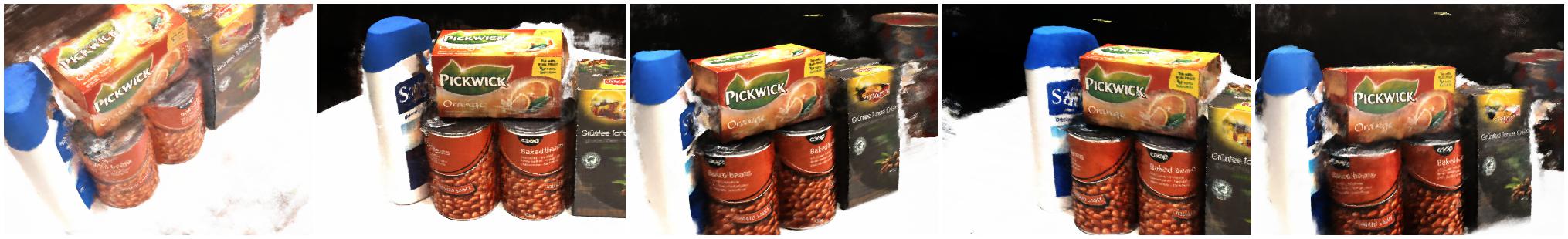}
        \vspace{-.52cm}
         \label{fig:supp_dtu_3_ours_2}
     \end{subfigure}
     \begin{subfigure}[b]{0.9\textwidth}
         \centering
         \vspace{-.07cm}
        \includegraphics[width=\linewidth]{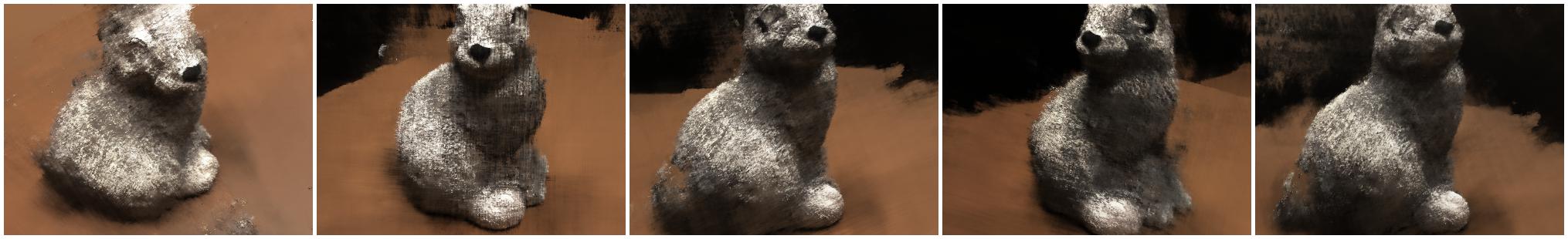}
        \vspace{-.52cm}
         \caption{\scriptsize 3-view}
         \label{fig:supp_dtu_3_ours_3}
     \end{subfigure}
     \begin{subfigure}[b]{0.9\textwidth}
         \centering
         \vspace{-.07cm}
        \includegraphics[width=\linewidth]{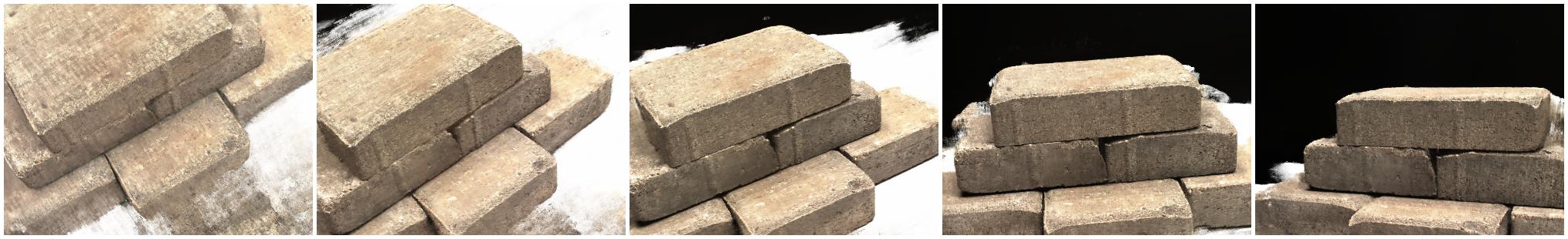}
        \vspace{-.52cm}
         \label{fig:supp_dtu_6_ours_1}
     \end{subfigure}
     \begin{subfigure}[b]{0.9\textwidth}
         \centering
         \vspace{-.07cm}
        \includegraphics[width=\linewidth]{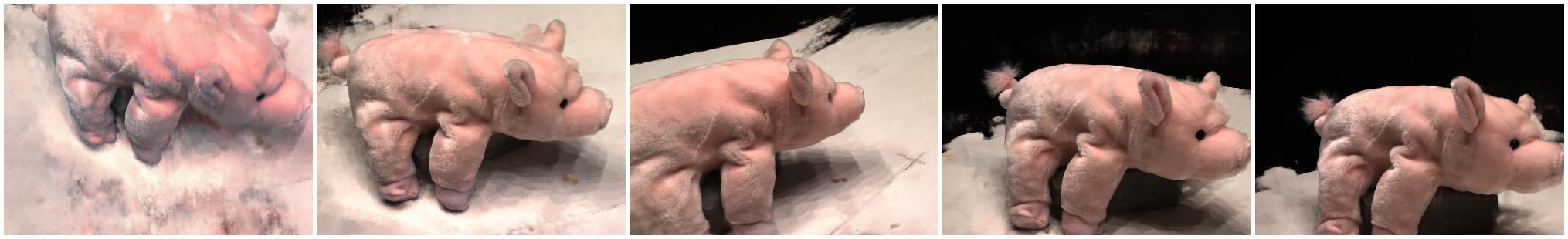}
        \vspace{-.52cm}
         \label{fig:supp_dtu_6_ours_2}
     \end{subfigure}
     \begin{subfigure}[b]{0.9\textwidth}
         \centering
         \vspace{-.07cm}
        \includegraphics[width=\linewidth]{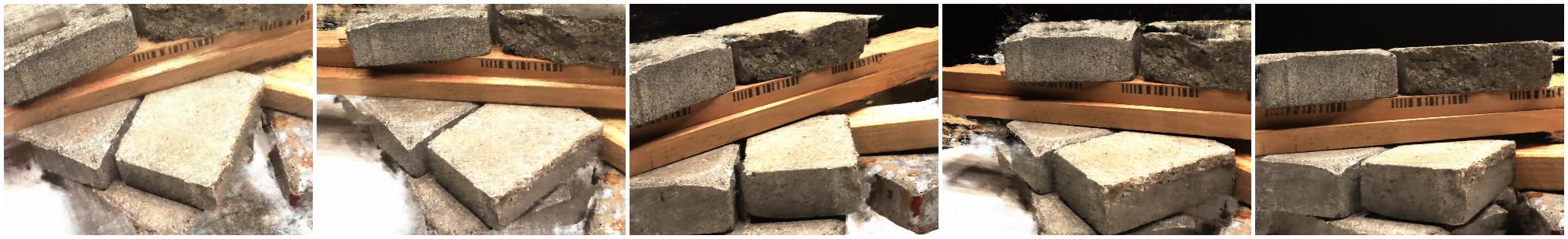}
        \vspace{-.52cm}
         \caption{\scriptsize 6-view}
         \label{fig:supp_dtu_6_ours_3}
     \end{subfigure}
     \begin{subfigure}[b]{0.9\textwidth}
         \centering
         \vspace{-.07cm}
        \includegraphics[width=\linewidth]{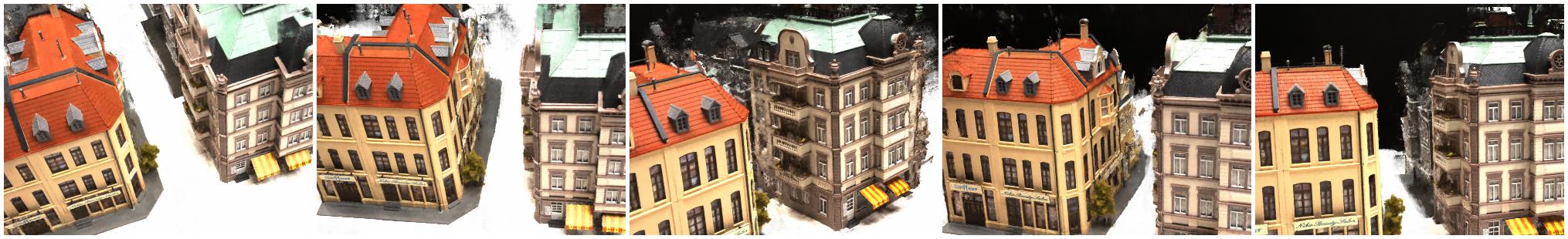}
        \vspace{-.52cm}
         \label{fig:supp_dtu_9_ours_1}
     \end{subfigure}
     \begin{subfigure}[b]{0.9\textwidth}
         \centering
         \vspace{-.07cm}
        \includegraphics[width=\linewidth]{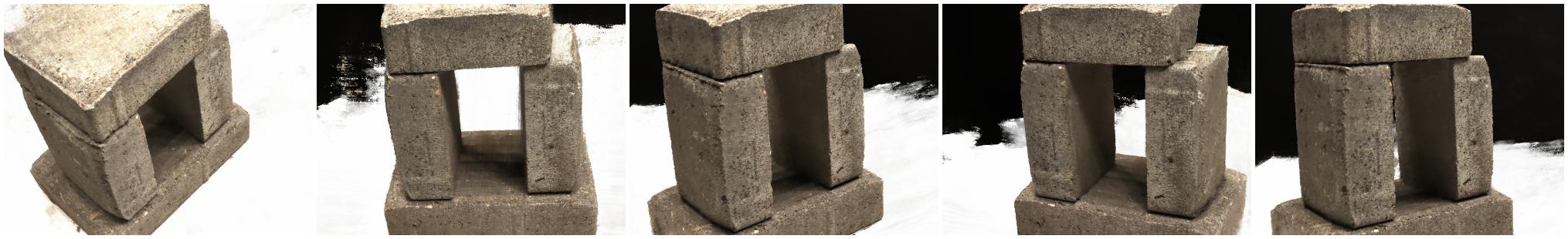}
        \vspace{-.52cm}
         \label{fig:supp_dtu_9_ours_2}
     \end{subfigure}
     \begin{subfigure}[b]{0.9\textwidth}
         \centering
         \vspace{-.07cm}
        \includegraphics[width=\linewidth]{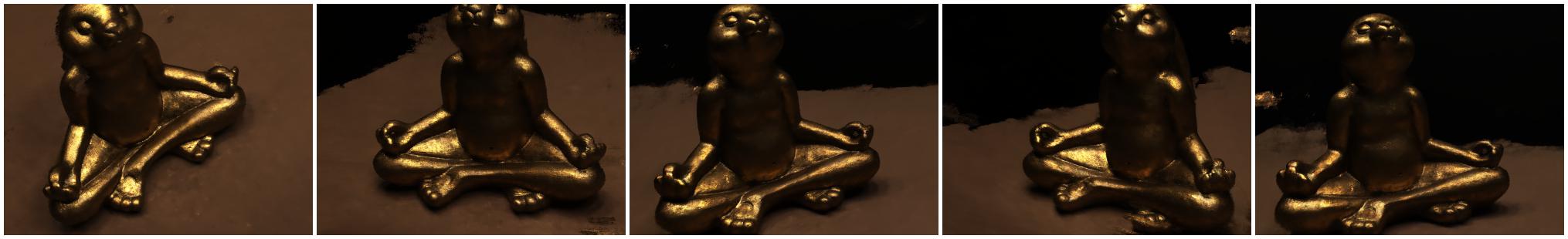}
        \vspace{-.52cm}
         \caption{\scriptsize 9-view}
         \label{fig:supp_dtu_9_ours_3}
     \end{subfigure}
     \vspace{-.25cm}
    \caption{
    \textbf{Additional qualitative results of our MixNeRF on DTU.}
    }
    \label{fig:supp_qual_res_dtu_ours}
\end{figure*}

\begin{figure*}
     \centering
     \begin{subfigure}[b]{\textwidth}
         \centering
         \vspace{-.07cm}
        \includegraphics[width=\linewidth]{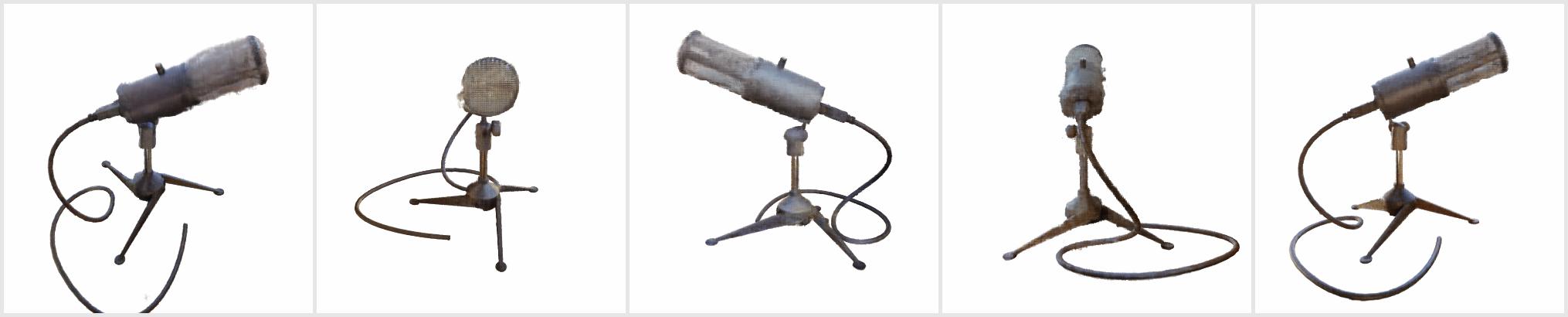}
        \vspace{-.52cm}
         \label{fig:supp_blender_4_ours_1}
     \end{subfigure}
     \begin{subfigure}[b]{\textwidth}
         \centering
         \vspace{-.07cm}
        \includegraphics[width=\linewidth]{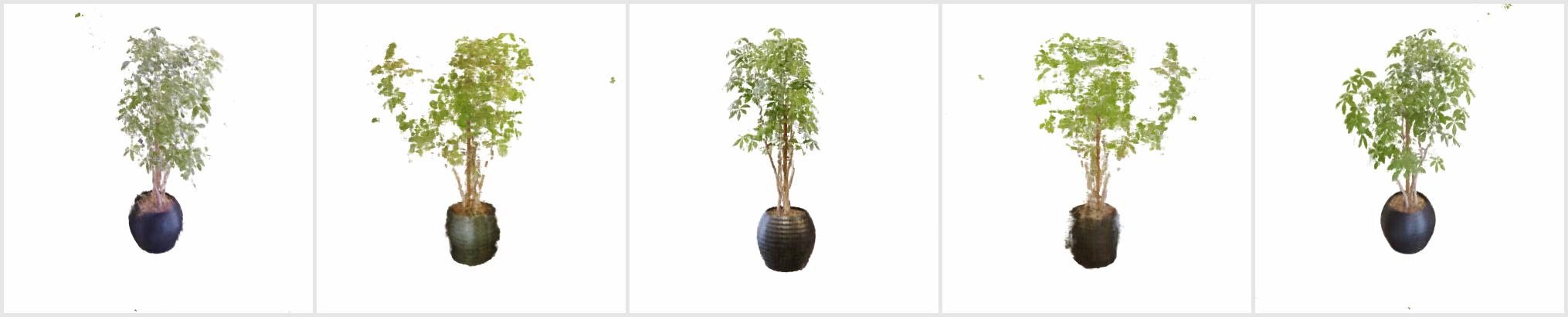}
        \vspace{-.52cm}
         \label{fig:supp_blender_4_ours_2}
     \end{subfigure}
     \begin{subfigure}[b]{\textwidth}
         \centering
         \vspace{-.07cm}
        \includegraphics[width=\linewidth]{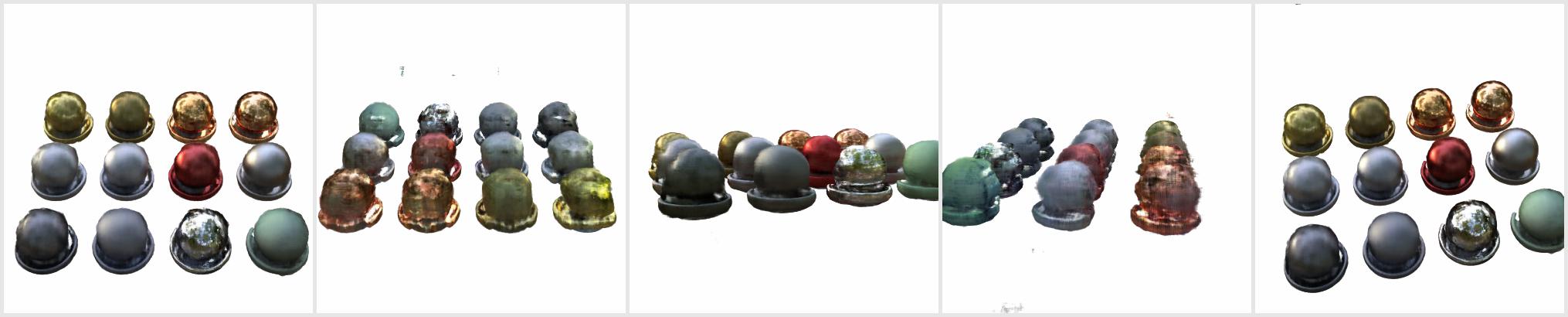}
        \vspace{-.52cm}
        \caption{\scriptsize 4-view}
         \label{fig:supp_blender_4_ours_3}
     \end{subfigure}
     \begin{subfigure}[b]{\textwidth}
         \centering
        \includegraphics[width=\linewidth]{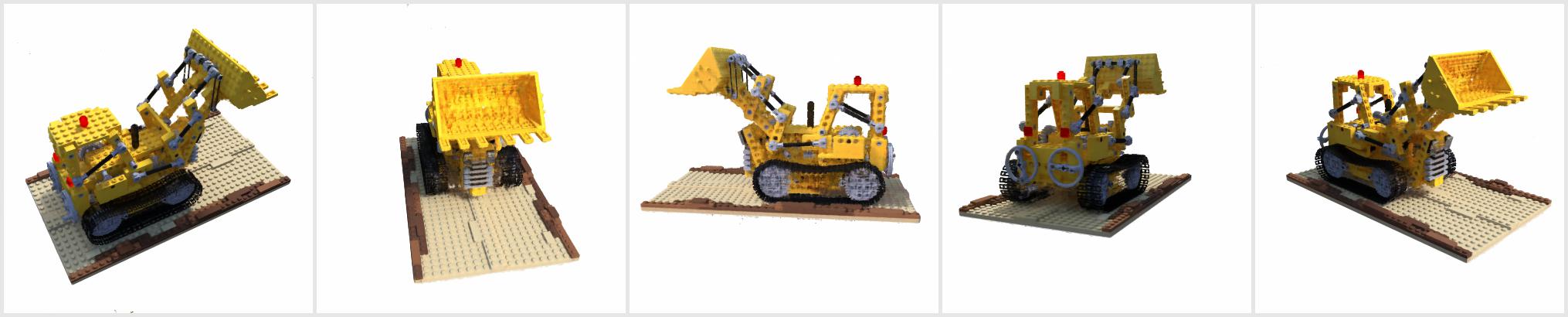}
        \vspace{-.52cm}
         \label{fig:supp_blender_8_ours_1}
     \end{subfigure}
     \begin{subfigure}[b]{\textwidth}
         \centering
         \vspace{-.07cm}
        \includegraphics[width=\linewidth]{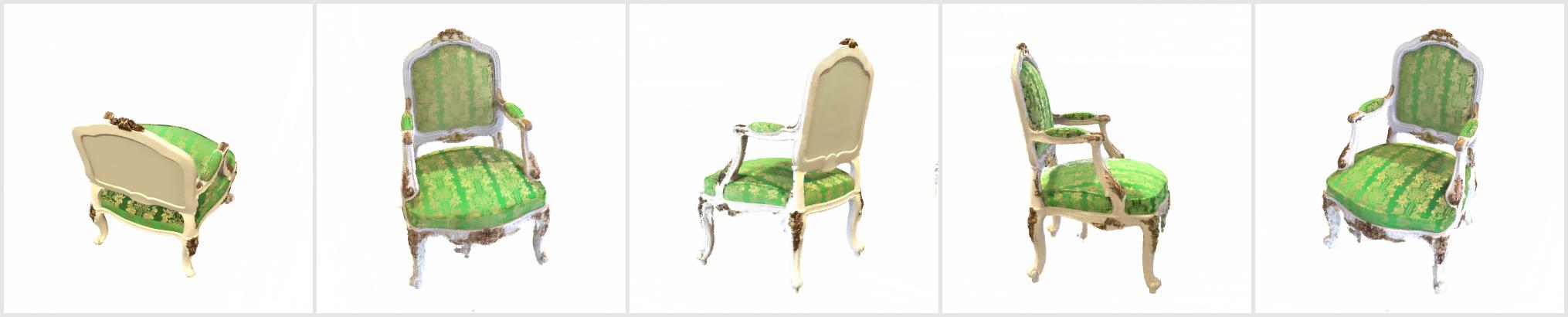}
        \vspace{-.52cm}
         \label{fig:supp_blender_8_ours_2}
     \end{subfigure}
     \begin{subfigure}[b]{\textwidth}
         \centering
         \vspace{-.07cm}
        \includegraphics[width=\linewidth]{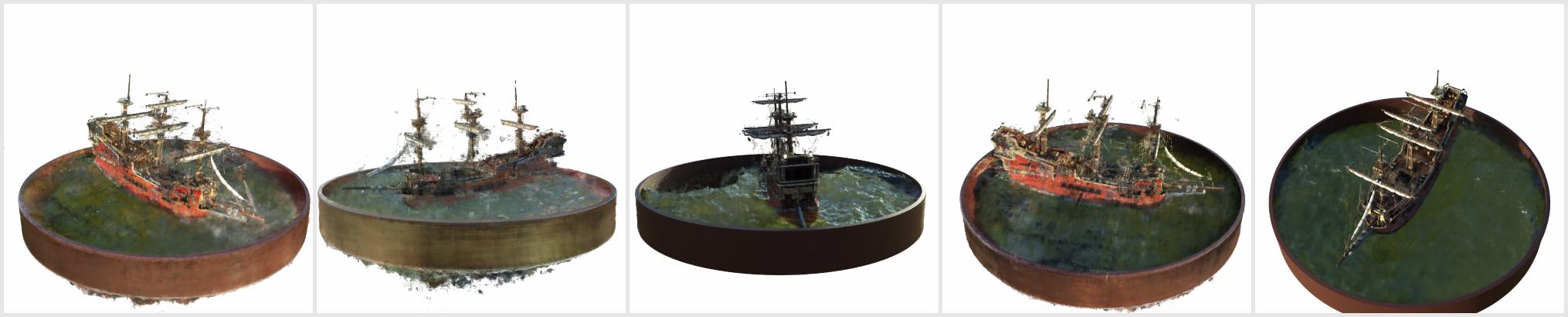}
        \vspace{-.52cm}
        \caption{\scriptsize 8-view}
         \label{fig:supp_blender_8_ours_3}
     \end{subfigure}
     \vspace{-.25cm}
    \caption{
    \textbf{Additional qualitative results of our MixNeRF on Realistic Synthetic 360$^\circ$.}
    }
    \label{fig:supp_qual_res_blender_ours}
\end{figure*}

We demonstrate the additional qualitative comparisons in \cref{fig:supp_qual_res_llff}, \cref{fig:supp_qual_res_dtu}, and \cref{fig:supp_qual_res_blender}.
Moreover, we show the additional qualitative results of our MixNeRF in \cref{fig:supp_qual_res_llff_ours}, \cref{fig:supp_qual_res_dtu_ours}, and \cref{fig:supp_qual_res_blender_ours}.

\section{Limitations and Future Work}
Our MixNeRF achieves the state-of-the-art performance without any extra training resources, \eg additional inference for pre-generated rays from unseen viewpoints, external modules for providing supplemental supervisory signals, or so on.
However, it still shows a few degenerate parts in the rendered images under the very sparse scenario as few as 3-view, due to the disturbance from the non-objects, \eg a background or a table, especially on the DTU dataset.
To eliminate the artifacts more effectively, developing an algorithm for classifying the pixels into an object or non-object can be a promising future work.

\section{Potential Negative Societal Impact}
Our method is able to synthesize a photo-realistic image from novel view from the limited training resources.
Although it provides much benefits for practical applications where the dense training resources are hard to collect, there exists a possibility of negative consequences with malicious intents, \eg a misleading content made with an intent to either conceal or show some specific views.
Therefore, the effort to prevent the malicious usage should be made, \eg strictly checking on the permission to use sensitive data, deep fake detection, and so on.

\end{document}